\documentclass[runningheads]{llncs}
\usepackage{graphicx}

\usepackage{tikz}
\usepackage{comment}
\usepackage{amsmath,amssymb} 
\usepackage{color}
\usepackage{epsfig}
\usepackage{graphicx}
\usepackage{amsmath}
\usepackage{amssymb}
\usepackage{calc}
\usepackage{tabularx}
\usepackage{adjustbox}
\usepackage{bm}
\usepackage{multirow}
\usepackage{booktabs}
\usepackage{amsmath,amsfonts,amssymb}
\usepackage{xcolor}

\usepackage{enumitem}
\usepackage{caption} 
\pdfoutput=1
\usepackage{hyperref}
\hypersetup{
    colorlinks=true,
    linkcolor=black,
    citecolor=purple,
    filecolor=black,
    urlcolor=black,
}

\newcommand{\NDCG}{NDCG$\uparrow$}
\newcommand{\MRR}{MRR$\uparrow$}
\newcommand{\ROne}{R@1$\uparrow$}
\newcommand{\RFive}{R@5$\uparrow$}
\newcommand{\RTen}{R@10$\uparrow$}
\newcommand{\Mean}{Mean$\downarrow$}

\begin{document}
\pagestyle{headings}
\mainmatter
\def\ECCVSubNumber{4553}

\title{Efficient Attention Mechanism for Visual Dialog that can Handle All the Interactions between Multiple Inputs}

\titlerunning{Efficient Attention Mechanism for Multiple Inputs in Visual Dialog}

\author{Van-Quang Nguyen\inst{1} \and
Masanori Suganuma\inst{2, 1} \and
Takayuki Okatani\inst{1, 2}}

\authorrunning{Van-Quang Nguyen et al.} 

\institute{Grad School of Information Sciences, 
Tohoku University 
\and RIKEN Center for AIP \\
\email{\{quang, suganuma, okatani\}@vision.is.tohoku.ac.jp}}

\maketitle

\begin{abstract}
It has been a primary concern in recent studies of vision and language tasks to design an effective attention mechanism dealing with interactions between the two modalities. The Transformer has recently been extended and applied to several bi-modal tasks, yielding promising results. For visual dialog, it becomes necessary to consider interactions between three or more inputs, i.e., an image, a question, and a dialog history, or even its individual dialog components. In this paper, we present a neural architecture named {\em Light-weight Transformer for Many Inputs} (LTMI) that can efficiently deal with all the interactions between multiple such inputs in visual dialog. It has a block structure similar to the Transformer and employs the same design of attention computation, whereas it has only a small number of parameters, yet has sufficient representational power for the purpose. Assuming a standard setting of visual dialog, a layer built upon the proposed attention block has less than one-tenth of parameters as compared with its counterpart, a natural Transformer extension. The experimental results on the VisDial datasets validate the effectiveness of the proposed approach, showing improvements of the best NDCG score on the VisDial v1.0 dataset from 57.59 to 60.92 with a single model, from 64.47 to 66.53 with ensemble models, and even to 74.88 with additional finetuning.
\end{abstract}

\section{Introduction} \label{introduction}
Recently, an increasing amount of attention has been paid to problems lying at the intersection of the vision and language domains. Many pilot tasks in this intersecting region have been designed and introduced to the research community, together with datasets.
Visual dialog has been developed aiming at a higher level of vision-language interactions \cite{das2017visual}, as compared with VQA (visual question answering) \cite{antol2015vqa} and VCR (visual commonsense reasoning). 
It extends VQA to multiple rounds; given an image and a history of question-answer pairs about the image, an agent is required to answer a new question. 
For example, to answer the question {\it `What color are they?'}, the agent needs to understand the context from a dialog history to know what {\it `they'} refers to and look at the relevant image region to find out a color.

In recent studies of vision-language tasks, a primary concern has been to design an attention mechanism that can effectively deal with interactions between the two modalities. In the case of visual dialog,  it becomes further necessary to consider interactions between an image, a question, and a dialog history or additionally multiple question-answer pairs in the history.
Thus, the key to success will be how to deal with such interactions between three and more entities. Following a recent study \cite{Schwartz2019FactorGA}, we will use the term {\em utility} to represent each of these input entities for clarity, since the term {\em modality} is inconvenient to distinguish between the question and the dialog history. 

Existing studies have considered attention from one utility to another based on different hypotheses, such as ``question $\rightarrow$ history $\rightarrow$ image" path in \cite{Kang2019DualAN,lu2017best}, and ``question $\rightarrow$ image $\rightarrow$ history $\rightarrow$ question" path in \cite{Gan2019MultistepRV,wu2018you}, etc. These methods cannot take all the interactions between utilities into account, although the missing interactions could be crucial. Motivated by this, a recent study tries to capture all the possible interactions by using a factor graph \cite{Schwartz2019FactorGA}. 
However, building the factor graph is computationally inefficient, which seemingly hinders the method from unleashing the full potential of modeling all the interactions, especially when the dialog history grows long.

The Transformer \cite{vaswani2017attention} has become a standard neural architecture for various tasks in the field of natural language processing, especially since the huge success of its pretrained model, BERT \cite{devlin2018bert}. Its basic mechanism has recently been extended to the bi-modal problems of vision and language, yielding promising results \cite{chen2019uniter,gao2019dynamic,li2019visualbert,lu2019vilbert,yu2019deep}. Then, it appears to be natural to extend it further to deal with many-to-many utility interactions. However, it is not easy due to several reasons.
As its basic structure is designed to be deal with self-attention, even in the simplest case of bi-modality, letting $X$ and $Y$ be the two utilities, there are four patterns of attention, $X\rightarrow Y$, $Y\rightarrow X$, $X\rightarrow X$, and $Y\rightarrow Y$; we need an independent Transformer block for each of these four. 
When extending this to deal with many-to-many utility interactions, the number of the blocks and thus of their total parameters increases proportionally with the square of the number of utilities, making it computationally expensive. Moreover, it is not apparent how to aggregate the results from all the interactions.

To cope with this, we propose a neural architecture named {\em Light-weight Transformer for Many Inputs} (LTMI) that can deal with all the interactions between many utilities. While it has a block structure similar to the Transformer and shares the core design of attention computation, it differs in the following two aspects. One is the difference in the implementation of multi-head attention. Multi-head attention in the Transformer projects the input feature space linearly to multiple lower-dimensional spaces, enabling to handle multiple attention maps, where the linear mappings are represented with learnable parameters. In the proposed model, we instead split the input feature space to subspaces mechanically according to its indexes, removing all the learnable parameters from the attention computation. 

The other difference from the Transformer is that LTMI is designed to receive multiple utilities and compute all the interactions to one utility from all the others including itself. This yields the same number of attended features as the input utilities, which are then concatenated in the direction of the feature space dimensions and then linearly projected back to the original feature space. We treat the parameters of the last linear projection 
as only learnable parameters in LTMI.
This design makes it possible to retain sufficient representational power with a much fewer number of parameters, as compared with a natural extension of the Transformer block to many utilities.
By using the same number of blocks in parallel as the number of utilities, we can deal with all the interactions between the utilities; see Fig.~\ref{fig:simplesymbol} for example.
Assuming three utilities and the feature space dimensionality of $512$, a layer consisting of
LTMI has 2.38M parameters, whereas its counterpart based on  naive Transformer extension has 28.4M parameters.

The contribution of this study is stated as follows. i) A novel attention mechanism for visual dialog that can deal with all interactions between multiple utilities is proposed. It is lightweight and yet has a sufficient representational power. ii) A series of experiments and ablative studies are conducted, achieving the new state-of-the-art results on the VisDial datasets, e.g., high NDCG scores on the VisDial v1.0 dataset. These validate the effectiveness of the proposed method. iii) 
The visualization of the inference conducted by our method is given, providing interpretation of how the proposed mechanism works. 
Our method achieves the third place with only a small gap to the first place in the Visual Dialog 2019 leaderboard; the result is achieved without using external training data and with a fewer number of parameters.

\section{Related Work}
\subsection{Attention Mechanisms for Vision-Language Tasks}

Attention mechanisms are currently indispensable to build neural architectures for vision-language tasks, such as VQA \cite{chen2015abc,ilievski2016focused,kim2018bilinear,lu2016hierarchical,nguyen2018improved,yang2016stacked,yu2017multi,yu2018beyond} and visual grounding \cite{deng2018visual,yu2018mattnet,zhuang2018parallel}, etc. Inspired by the recent success of the Transformer for language tasks \cite{devlin2018bert,vaswani2017attention}, several studies have proposed its extensions to bi-modal vision-language tasks \cite{chen2019uniter,gao2019dynamic,li2019visualbert,lu2019vilbert,tan2019lxmert,yu2019deep}. Specifically, for VQA, it is proposed to use intra-modal and inter-modal attention blocks and stack them alternately to fuse question and image features \cite{gao2019dynamic}; it is also proposed to use a cascade of modular co-attention layers that compute the self-attention and guided-attention of question and image features \cite{yu2019deep}. The method of pretraining a Transformer model used in BERT \cite{devlin2018bert} is employed along with Transformer extension to bi-modal tasks for several vision-language tasks \cite{chen2019uniter,li2019visualbert,lu2019vilbert}. They first pretrain the models on external datasets, such as COCO Captions \cite{chen2015microsoft} or Conceptual Captions dataset \cite{sharma2018conceptual}, and then fine-tune them on several target tasks.

\subsection{Visual Dialog}
The task of visual dialog has recently been proposed by two groups of researchers concurrently \cite{das2017visual,de2017guesswhat}. De Vries et al. introduced the GuessWhat?! dataset, which is built upon goal-oriented dialogs held by two agents to identify unknown objects in an image through a set of yes/no questions \cite{de2017guesswhat}. Das et al. released the VisDial dataset, which is built upon dialogs consisting of pairs of a question and an answer about an image that are provided in the form of natural language texts  \cite{das2017visual}. Kottur et al. recently introduced CLEVR-Dialog as the diagnostic dataset for visual dialog \cite{kottur2019clevr}.

Most of the existing approaches employ an encoder-decoder architecture \cite{sutskever2014sequence}. They can be categorized into the following three groups by the design of the encoder: i) fusion-based methods, e.g., LF \cite{das2017visual} and HRE \cite{das2017visual}, which fuses the inputs by their concatenation followed by the application of a feed-forward or recurrent network, and Synergistic \cite{Guo_2019_CVPR}, which fuses the inputs at multiple stages;
ii) attention-based methods that compute attended features of the input image, question, and history utilities, e.g., MN \cite{das2017visual}, CoAtt \cite{wu2018you}, HCIAE \cite{lu2017best}, Synergistic \cite{Guo_2019_CVPR}, ReDAN \cite{Gan2019MultistepRV}, FGA \cite{Schwartz2019FactorGA}, and CDF \cite{kim2020modality}; ReDAN compute the attention over several reasoning steps, FGA models all the interactions over many utilities via a factor graph; 
iii) methods that attempt to resolve visual co-reference, e.g., RvA \cite{Niu_2019_CVPR} and  CorefNMN \cite{kottur2018visual}, which use neural modules to form an attention mechanism, DAN \cite{Kang2019DualAN}, which employs a network having two attention modules, and AMEM \cite{seo2017visual}, which utilizes a  memory mechanism for attention. As for the decoder, there are two designs: i) discriminative decoders that rank the candidate answers using the cross-entropy loss \cite{das2017visual} or the n-pair loss \cite{lu2017best}; and ii) generative decoders that yield an answer by using a MLE loss \cite{das2017visual}, weighted likelihood estimation \cite{Zhang2019}, or a combination with adversarial learning \cite{lu2017best,wu2018you}, which trains a discriminator on both positive and negative answers, then transferring it to the generator with auxiliary adversarial learning.

Other approaches include GNN \cite{Zheng2019ReasoningVD}, which models relations in a dialog by an unknown graph structure;
the employment of reinforcement learning \cite{chattopadhyay2017evaluating,das2017learning}; 
and HACAN \cite{Yang2019MakingHM} which adopts policy gradient to learn the impact of history by intentionally imposing the wrong answer into dialog history. 
In \cite{wang2020vd,murahari2019large}, pretrained vision-language models 
are adopted, which consist of many Transformer blocks with hundreds of millions parameters, leading to some performance gain.
Qi et al. \cite{qi2020two} present model-agnostic principles for visual dialog to maximize performance. 

\section{Efficient Attention Mechanism for Many Utilities}

\subsection{Attention Mechanism of Transformer}
As mentioned earlier, the Transformer has been applied to several bi-modal vision-language tasks, yielding promising results.
The Transformer computes and uses attention from three types of inputs, $Q$ (query), $K$ (key), and $V$ (value). Its computation is given by
\begin{equation}
    {\cal A}(Q,K,V)=\mbox{softmax}\left(\frac{Q K^\top}{\sqrt{d}} \right) V,
    \label{eqn:trm_attn}
\end{equation}
where $Q$, $K$, and $V$ are all collection of features, each of which is represented by a $d$-dimensional vector. To be specific, $Q=[q_1,\ldots,q_M]^\top\in \mathbb{R}^{M\times d}$ is a collection of $M$ features; similarly, $K$ and $V$ are each a collection of $N$ features, i.e.,  $K, V\in \mathbb{R}^{N\times d}$. In Eq.(\ref{eqn:trm_attn}), $V$ is attended with the weights computed from the similarity between $Q$ and $K$.

The above computation is usually multi-plexed in the way called multi-head attention. It enables to use a number of attention distributions in parallel, aiming at an increase in representational power. The outputs of $H$ `heads' are concatenated, followed by linear transformation with learnable weights $W^O\in\mathbb{R}^{d\times d}$ as
\begin{equation}
    {\cal A}^{\mathrm{M}}(Q,K,V)=\begin{bmatrix}
    \mathrm{head}_1,\cdots,\mathrm{head}_H
    \end{bmatrix}W^O.
\end{equation}
Each head is computed as follows:
\begin{equation}
    \mathrm{head}_h = {\cal A}(QW_h^Q, KW_h^K, VW_h^V), \;\;h=1,\ldots,H, 
\end{equation}
where $W_h^Q$, $W_h^K$, $W_h^V \in \mathbb{R}^{d\times d_H}$ each are learnable weights inducing a linear projection from the feature space of $d$-dimensions to a lower space of $d_H(=d/H)$-dimensions. Thus, one attentional block ${\cal A}^{\mathrm{M}}(Q,K,V)$ has the following learnable weights:
\begin{equation}
    (W_1^Q, W_1^K, W_1^V),\cdots,(W_H^Q, W_H^K, W_H^V)\;\; \mbox{and} \;\; W^O.
    \label{eqn:ma_weights}
\end{equation}

\begin{figure}[t]
\centering
\includegraphics[width=0.65\columnwidth]{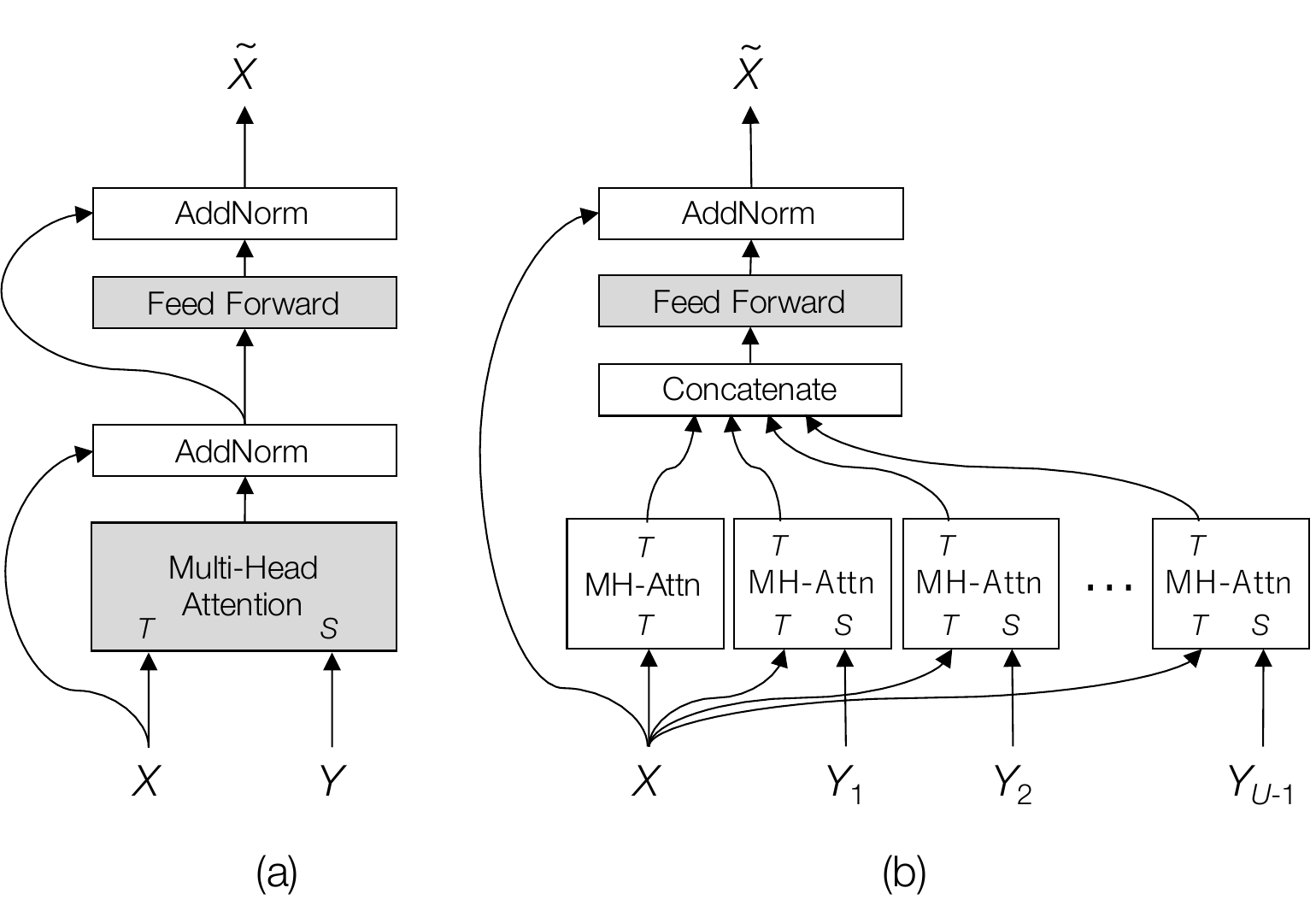}
\vspace*{-0.1in}
\caption{(a) Source-to-target attention for bi-modal problems implemented by the standard Transformer block; the source $Y$ is attended by weights computed from the similarity between the target $X$ and $Y$. (b) The proposed block that can deal with many utilities; the source features $\{Y_1,\ldots,Y_{U-1}\}$ are attended by weights computed between them and the target $X$. Shaded boxes have learnable weights }
\label{fig:comparison_models}
\end{figure}

\subsection{Application to Bi-Modal Tasks}
While $Q$, $K$, and $V$ in NLP tasks are of the same modality (i.e., language), the above mechanism has been extended to bi-modality and applied to vision-language tasks in recent studies \cite{chen2019uniter,gao2019dynamic,li2019visualbert,lu2019vilbert,tan2019lxmert,yu2019deep}. 
They follow the original idea of the Transformer, considering attention from source features $Y$ to target features $X$ as
\begin{equation}\label{eqn:basic_attn}
   {\cal A}_Y(X) = {\cal A}^{\mathrm{M}}(X, Y, Y).
\end{equation}
In MCAN \cite{yu2019deep}, language feature is treated as the source and visual feature is as the target. 
In \cite{li2019visualbert} and others \cite{chen2019uniter,gao2019dynamic,lu2019vilbert,tan2019lxmert}, co-attention, i.e., attention in the both directions, is considered. Self-attention, i.e., the attention from features to themselves, is given as a special case by 
\begin{equation}
    {\cal A}_X(X) = {\cal A}^{\mathrm{M}}(X, X, X).
    \label{eqn:a_xx_attn}
\end{equation}
In the above studies, the Transformer block with the source-to-target attention and that with the self-attention are independently treated and are stacked, e.g., alternately or sequentially.

\subsection{Light-weight Transformer for Many Inputs} 

Now suppose we wish to extend the above attention mechanism to a greater number of utilities\footnote{As we stated in Introduction, we use the term {\em utility} here to mean a collection of features.};
we denote the number by $U$. If we consider every possible source-target pairs, 
there are $U(U-1)$ cases in total, as there are $U$ targets, for each of which $U-1$ sources exist. Then we need to consider attention computation ${\cal A}_{Y}(X)$ over $U-1$ sources $Y$'s for each target $X$.
Thus, the straightforward extension of the above attention mechanism to $U$ utilities needs $U(U-1)$ times the number of parameters listed in Eq.(\ref{eqn:ma_weights}). If we stack the blocks, the total number of parameters further increases proportionally.

To cope with this, we remove all the weights from Eq.(\ref{eqn:basic_attn}). To be specific, for each head $h(=1,\ldots,H)$, we choose and freeze $(W_h^Q, W_h^K, W_h^V)$ as
\begin{equation}
W^Q_h=W^K_h=W^V_h = [\underbrace{O_{d_H},\cdots,O_{d_H}}_{(h-1)d_H},I_{d_H},\underbrace{O_{d_H},\cdots,O_{d_H}}_{(H-h)d_H}]^\top,
    \label{eqn:freeze_weights}
\end{equation}
where $O_{d_H}$ is a $d_H\times d_H$ zero matrix and $I_{d_H}$ is a $d_H\times d_H$ identity matrix. 
In short, the subspace for each head is determined to be one of $H$ subspaces obtained by splitting the $d$-dimensional feature space with its axis indexes. 
Besides, we set $W^O=I$, which is the linear mapping applied to the concatenation of the heads' outputs. 
Let $\bar{\cal A}_Y(X)$ denote this simplified attention mechanism.

Now let the utilities be denoted by $\{X,Y_1,\ldots,Y_{U-1}\}$, where $X\in \mathbb{R}^{M\times d}$ is the chosen target and others  $Y_i\in \mathbb{R}^{N_i\times d}$ are the sources. Then, we compute all the source-to-target attention as $\bar{\cal A}_{Y_1}(X),\cdots, \bar{\cal A}_{Y_{U-1}}(X)$. In the standard Transformer block (or rigorously its natural extensions to bi-modal problems), the attended features are simply added to the target as $X + {\cal A}_Y(X)$, followed by normalization and subsequent computations. To recover some of the loss in representational power due to the simplification yielding $\bar{\cal A}_Y(X)$, we propose a different approach to aggregate $\bar{\cal A}_{Y_1}(X),\cdots, \bar{\cal A}_{Y_{U-1}}(X)$ and $X$. 
Specifically, we concatenate all the source-to-target attention plus the self-attention $\bar{\cal A}_{X}(X)$ from $X$ to $X$ as
\begin{equation}
    X_{\mathrm{concat}} = [\bar{\cal A}_{X}(X), \bar{\cal A}_{Y_1}(X), \cdots,\bar{\cal A}_{Y_{U-1}}(X)],
\end{equation}
where $X_{\mathrm{concat}}\in\mathbb{R}^{M\times Ud}$. We then apply linear transformation to it given by 
$W\in \mathbb{R}^{Ud\times d}$ and $b\in \mathbb{R}^d$ with a single fully-connected layer, followed by the addition of the original $X$ and layer normalization as
\begin{equation}
    \tilde{X} = \mathrm{LayerNorm}( \mathrm{ReLU}(X_{\mathrm{concat}} W 
    +\mathbf{1}_{M} \cdot b^\top) + X),
\end{equation}
where $\mathbf{1}_M$ is  $M$-vector with all ones. With this method, we aim at recovery of representational power as well as the effective aggregation of information from all the utilities. 

\begin{figure}[t]
\centering
\includegraphics[width=0.65\columnwidth]{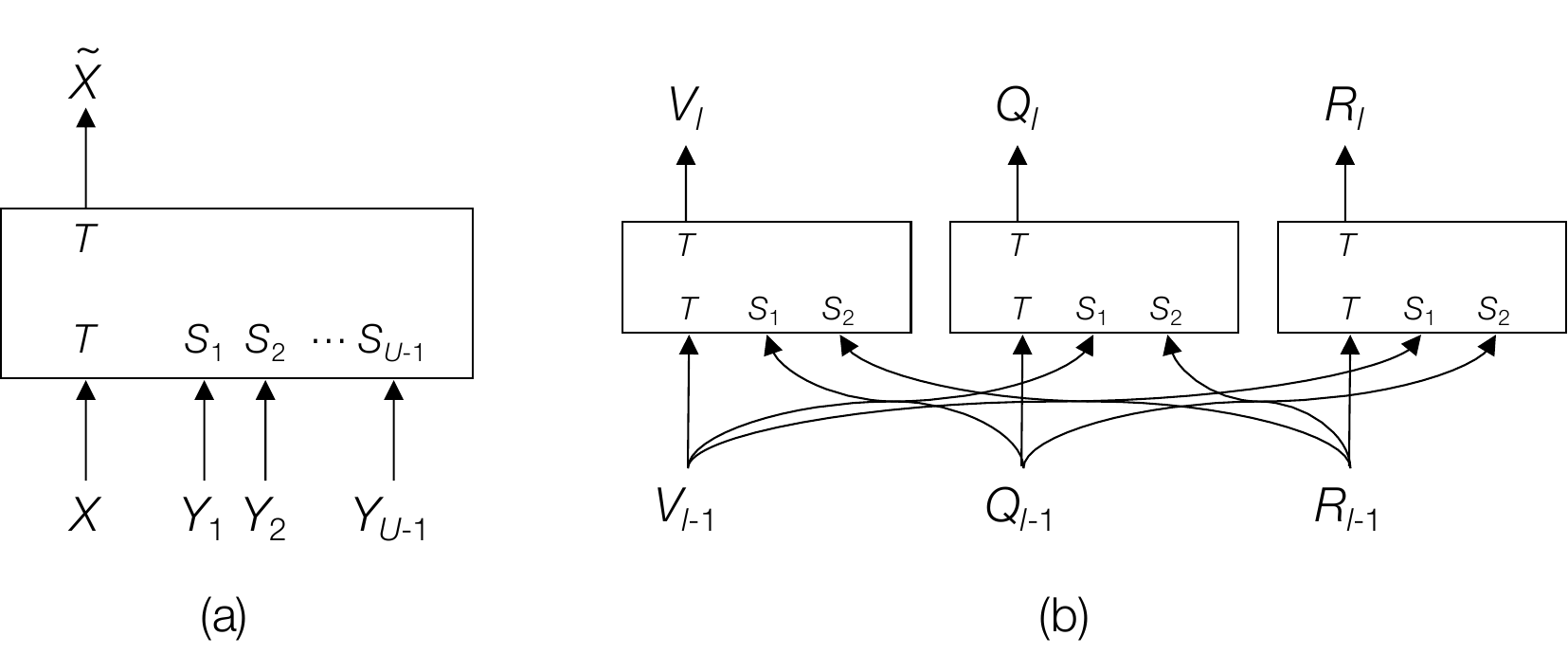} 
\vspace*{-0.1in}
\caption{(a) Simplified symbol of the proposed block shown in Fig.~\ref{fig:comparison_models}(b). (b) Its application to Visual Dialog }
\label{fig:simplesymbol}
\end{figure}

\subsection{Interactions between All Utilities}
We have designed a basic block (Fig.~\ref{fig:comparison_models}(b)) that deals with attention from many sources to a single target. We wish to consider all possible interactions between all the utilities, not a single utility being the only target. To do this, we use $U$ basic blocks to consider all the source-to-target attention. Using the basic block as a building block, we show how an architecture is designed for visual dialog having three utilities, visual features $V$, question features $Q$, and dialog history features $R$, in Fig.~\ref{fig:simplesymbol}(b).

\section{Implementation Details for Visual Dialog}

\subsection{Problem Definition}
The problem of Visual Dialog is stated as follows. An agent is given the image of a scene and a dialog history containing $T$ entities, which consists of a caption and question-answer pairs at $T-1$ rounds. Then, the agent is further given a new question at round $T$ along with 100 candidate answers for it and requested to answer the question by choosing one or scoring each of the candidate answers.

\subsection{Representation of Utilities}
We first extract features from an input image, a dialog history, and a new question at round $T$ to obtain their representations. For this, we follow the standard method employed in many recent studies. For the image utility, we use the bottom-up mechanism \cite{anderson2018bottom}, which extracts region-level image features using the Faster-RCNN \cite{ren2015faster} pre-trained on the Visual Genome dataset \cite{krishna2017visual}. For each region (i.e., a bounding box = an object), we combine its CNN feature and geometry to get a $d$-dimensional vector $v_i$ ($i=1,\ldots,K$), where $K$ is the predefined number of regions. We then define $V = [v_1, v_2, \cdots, v_K]^\top \in \mathbb{R}^{K \times d}$. For the question utility, after embedding each word using an embedding layer initialized by pretrained GloVe vectors, we use two-layer Bi-LSTM to transform them to $q_i$ $(i=1,\ldots,N)$, where $N$ is the number of words in the question. We optionally use the positional embedding widely used in NLP studies. We examine its effects in an ablation test. We then define $Q = [q_1,\ldots,q_N]^\top \in \mathbb{R}^{N \times d}$. For the dialog history utility, we choose to represent it as a single utility here. Thus, each of its entities represents the initial caption or the question-answer pair at one round. As with the question utility, we use the same embedding layer and a two-layer Bi-LSTM together with the positional embeddings for the order of dialog rounds to encode them with a slight difference in formation of an entity vector $r_i$ ($i=1,\ldots,T)$, where $T$ is the number of Q\&A plus the caption. We then define $R = [r_1,\ldots,r_T]^\top \in \mathbb{R}^{T \times d}$. More details are provided in the supplementary material.

\subsection{Overall Network Design}
\begin{figure}[t]
\centering
\includegraphics[width=1.0\textwidth]{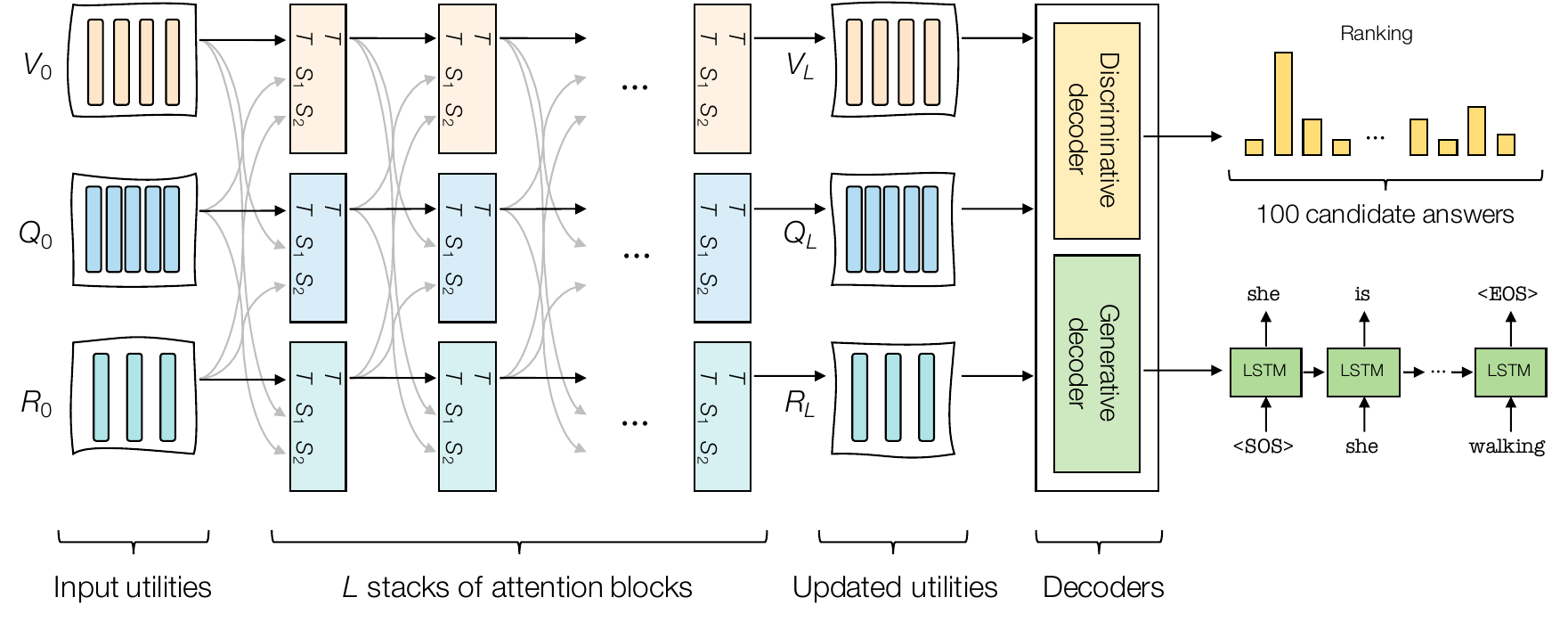}
\caption{The entire network built upon the proposed LTMI for Visual Dialog}
\label{fig:model_overview}
\vskip -0.1in
\end{figure}

Figure \ref{fig:model_overview} shows the entire network. It consists of an encoder and a decoder. The encoder consists of $L$ stacks of the proposed attention blocks; a single stack has $U$ blocks in parallel, as shown in Fig.~\ref{fig:simplesymbol}(b).
We set $V_0 = V$, $Q_0 = Q$, and $R_0 = R$ as the inputs of the first stack. After the $l$-th stack, the representations of the image, question, and dialog history utilities are updated as $V_l$, $Q_l$, and $R_l$, respectively.
In the experiments, we apply dropout with the rate of $0.1$ to the linear layer inside every block. There is a decoder(s) on top of the encoder. We consider a discriminative decoder and a  generative decoder, as in previous studies. Their design is explained below.

\subsection{Design of Decoders}
Decoders receive the updated utility representations, $V_L$, $Q_L$, and $R_L$ at their inputs. We convert them independently into $d$-dimensional vectors $c_V$, $c_Q$, and $c_R$, respectively. This conversion is performed by a simple self-attention computation. We take $c_V$ as an example here. First, attention weights over the entities of $V_L$ are computed by a two-layer network as 
\begin{equation}\label{eqn:a_v_weights}
    a_V = \mathrm{softmax}(\mathrm{ReLU}(V_LW_1 + \mathbf{1}_Kb_1^\top)W_2 + \mathbf{1}_Kb_2),
\end{equation}
where $W_1 \in \mathbb{R}^{d\times d}$, $W_2 \in \mathbb{R}^{d\times1}$, $b_1\in \mathbb{R}^d$, $b_2 \in \mathbb{R}^1$, and $\mathbf{1}_K$ is $K$-vector with all ones.
Then, $c_V$ is given by 
\begin{equation} \label{eqn:self_attn}
     c_V = \sum_{i = 1}^{K}v_{L,i}^\top a_{V,i},
\end{equation}
where $v_{L,i}$ is the $i$-th row vector of $V_L$ and $a_{V,i}$ is the $i$-th attention weight (a scalar). The others, i.e., $c_Q$ and $c_R$, can be obtained similarly. 

These vectors are integrated and used by the decoders. In our implementation for visual dialog, we found that $c_R$ does not contribute to better results; thus we use only $c_V$ and $c_Q$. Note that this does not mean the dialog utility $R$ is not necessary; it is interacted with other utilities inside the attention computation, contributing to the final prediction. The two $d$-vectors $c_V$ and $c_Q$ are concatenated as $[c_V^\top, c_Q^\top]^\top$, and this is projected to $d$-dimensional space, yielding a context vector $c\in \mathbb{R}^d$. 

We design the discriminative and generative decoders following the previous studies. Receiving $c$ and the candidate answers, the two decoders compute the score of each candidate answer in different ways. See details in the supplementary material. 

\subsection{Multi-Task Learning}

We observe in our experiments that accuracy is improved by training the entire network using the two decoders simultaneously. This is simply done by minimizing the sum of the losses, $\mathcal{L}_D$ for the discriminative one and $\mathcal{L}_G$ for the generative one (we do not use weights on the losses): 
\begin{equation}
    \mathcal{L} = \mathcal{L}_D + \mathcal{L}_G.
\end{equation}
The increase in performance may be attributable to the synergy of learning two tasks while sharing the same encoder. Details will be given in Sec.~\ref{sec:abalation_study}.

\section{Experimental Results}

\subsection{Experimental Setup}

\noindent
{\bf Dataset}~
We use the VisDial v1.0 dataset in our experiments which consists of the train 1.0 split (123,287 images), the val 1.0 split (2,064 images), and test v1.0 split (8,000 images).
Each image has a dialog composed of 10 question-answer pairs along with a caption. For each question-answer pair, 100 candidate answers are given.
The val v1.0 split and 2,000 images of the train v1.0 split are provided with dense annotations (i.e., relevance scores) for all candidate answers. Although the test v1.0 split was also densely annotated, the information about the ground truth answers and the dense annotations are not publicly available. Additionally, we evaulate the method on the Audio Visual Scene-aware Dialog Dataset \cite{hori2019end}; the results are shown in the supplementary.

\bigskip
\noindent
{\bf Evaluation metrics}~
From the visual dialog challenge 2018, normalized discounted cumulative gain (NDCG) has been used as the principal metric to evaluate methods on the VisDial v1.0 dataset. Unlike other classical retrieval metrics such as R@1, R@5, R@10, mean reciprocal rank (MRR), and mean rank, which are only based on a single ground truth answer, NDCG is computed based on the relevance scores of all candidate answers for each question, which can properly handle the case where each question has more than one correct answer, such as {\it `yes it is'} and {\it `yes'}; such cases do occur frequently. 

\bigskip
\noindent
{\bf Other configurations}~ We employ the standard method used by many recent studies for the determination of hyperparameters etc. For the visual features, we detect $K=100$ objects from each image. For the question and history features, we first build the vocabulary composed of 11,322 words that appear at least five times in the training split. The captions, questions, and answers are truncated or padded to 40, 20, and 20 words, respectively. Thus, $N=20$ for the question utility $Q$. $T$ for the history utilities varies depending on the number of dialogs. We use pre-trained 300-dimensional GloVe vectors \cite{pennington2014glove} to initialize the embedding layer, which is shared for all the captions, questions, and answers.

For the attention blocks, we set the dimension of the feature space to $d=512$ and the number of heads $H$ in each attention block to $4$. We mainly use models having two stacks of the proposed attention block. 
We train our models on the VisDial v0.9 and VisDial v1.0 dataset using the Adam optimizer \cite{kingma2014adam} with $5$ epochs and $15$ epochs respectively. The learning rate is warmed up from $1\times 10^{-5}$ to $1\times 10^{-3}$ in the first epoch, then halved every $2$ epochs. The batch size is set to $32$ for the both datasets.
\subsection{Comparison with State-of-the-art Methods}
\noindent
{\bf Compared methods}~
We compare our method with previously published methods on the VisDial v0.9 and VisDial v1.0 datasets, including LF, HRE, MN \cite{das2017visual}, LF-Att, MN-Att (with attention) \cite{das2017visual}, SAN \cite{yang2016stacked}, AMEM \cite{seo2017visual}, SF \cite{jain2018two}, HCIAE \cite{lu2017best} and Sequential CoAttention model (CoAtt) \cite{wu2018you}, Synergistic \cite{Guo_2019_CVPR}, FGA \cite{Schwartz2019FactorGA}, GNN \cite{Zheng2019ReasoningVD}, RvA \cite{Niu_2019_CVPR}, CorefNMN \cite{kottur2018visual}, DAN \cite{Kang2019DualAN}, and ReDAN \cite{Gan2019MultistepRV}, all of which were trained without using external datasets or data imposition.
Unless noted otherwise, the results of our models are obtained from the output of discriminative decoders.

\bigskip
\noindent{\bf
Results on the val v1.0 split}~
\begin{table}[t!]
\caption{Comparison of the performances of different methods on the validation set of VisDial v1.0 with discriminative and generative decoders.}
\resizebox{1.0\columnwidth}{!}{
\begin{small}
\begin{tabular}{lrccccccccccccc}
  \toprule
  & \multicolumn{1}{c}{\multirow{2}{*}{Model}} & \multicolumn{6}{c}{Discriminative}                    & & \multicolumn{6}{c}{Generative}                         \\
  \cmidrule{3-8} \cmidrule{10-15}
  & \multicolumn{1}{c}{}                       & \NDCG          & \MRR  & \ROne & \RFive& \RTen &\Mean & & \NDCG          & \MRR  & \ROne & \RFive& \RTen & \Mean \\
  \cmidrule{1-8} \cmidrule{10-15}
  & MN  \cite{das2017visual}                   & 55.13          & 60.42 & 46.09 & 78.14 & 88.05 & 4.63 & & 56.99          & 47.83 & 38.01 & 57.49 & 64.08 & 18.76 \\
  & CoAtt  \cite{wu2018you}                    & 57.72          & 62.91 & 48.86 & 80.41 & 89.83 & 4.21 & & 59.24          & 49.64 & 40.09 & 59.37 & 65.92 & 17.86 \\
  & HCIAE  \cite{lu2017best}                   & 57.75          & 62.96 & 48.94 & 80.5  & 89.66 & 4.24 & & 59.70          & 49.07 & 39.72 & 58.23 & 64.73 & 18.43 \\
  & ReDAN  \cite{Gan2019MultistepRV}           & 59.32          & 64.21 & 50.6  & 81.39 & 90.26 & 4.05 & & 60.47          & 50.02 & 40.27 & 59.93 & 66.78 & 17.4  \\
  \midrule
  & \textbf{LTMI}                                       & \textbf{62.72} & 62.32 & 48.94 & 78.65 & 87.88 & 4.86 & & \textbf{63.58} & 50.74 & 40.44 & 61.61 & 69.71 & 14.93 \\
  \bottomrule
\end{tabular}
\end{small}
}
\label{tab:table_val_v1}
\end{table}
We first compare single-model performance on the val v1.0 split. 
We select here MN, CoAtt, HCIAE, and ReDAN for comparison, as their performances from the both decoders in all metrics are available in the literature. To be specific, we use the accuracy values reported in \cite{Gan2019MultistepRV} for a fair comparison, in which these methods are reimplemented using the bottom-up-attention features.
Similar to ours, all these methods employ the standard design of discriminative and generative decoders as in \cite{das2017visual}. Table \ref{tab:table_val_v1} shows the results. It is seen that our method outperforms all the compared methods on the NDCG metric with large margins regardless of the decoder type. Specifically, as compared with ReDAN, the current state-of-the-art on the VisDial v1.0 dataset, our model has improved NDCG from 59.32 to 62.72 and from 60.47 to 63.58 with discriminative and generative decoders, respectively. 

\bigskip
\noindent{\bf Results on the test-standard v1.0 split}~
We next consider performance on the test-standard v1.0 split. In our experiments, we encountered a phenomenon that accuracy values measured by NDCG and other metrics show a trade-off relation (see the supplementary material for details), depending much on the choice of metrics (i.e., NDCG or others) for judging convergence at the training time. This is observed in the results reported in \cite{Gan2019MultistepRV} and is attributable to the inconsistency between the two types of metrics. Thus, we show two results here, the one obtained using NDCG for judging convergence and the one using MRR for it; the latter is equivalent to performing early stopping.

Table \ref{tab:result_visdial_1.0}(a) shows single-model performances on the blind test-standard v1.0 split. With the outputs from the discriminative decoder, our model gains improvement of 3.33pp in NDCG from the best model. When employing the aforementioned early stopping, our model achieves at least comparable or better performance in other metrics as well.

\begin{table}[t]
\caption{Comparison in terms of (a) single- and (b) ensemble-model performance on the blind test-standard v1.0 split of the VisDial v1.0 dataset and in terms of (c) the number of parameters of the attention mechanism. The result obtained by early stopping on MRR metric is denoted by $\star$ and those with fine-tuning on dense annotations are denoted by $\dagger$.}

\vspace*{-0.15in}
\begin{center}
\resizebox{1.0\columnwidth}{!}{
\begin{tabular}{ccc}
    \raisebox{0ex - \height}{
    \begin{tabular}{rcccccc}
    \multicolumn{7}{c}{\large \centering a) Performance of single models}\\
      \toprule 
      Model & \textbf{NDCG} $\uparrow $ & MRR $\uparrow$  & R@1 $\uparrow$    & R@5 $\uparrow$  & R@10 $\uparrow$  & Mean $\downarrow$ \\
      \midrule 
        LF \cite{das2017visual}            & 45.31 & 55.42 & 40.95 & 72.45 & 82.83 & 5.95 \\
        HRE \cite{das2017visual}           & 45.46 & 54.16 & 39.93 & 70.45 & 81.50 & 6.41 \\
        MN \cite{das2017visual}            & 47.50 & 55.49 & 40.98 & 72.30 & 83.30 & 5.92 \\
        MN-Att \cite{das2017visual}        & 49.58 & 56.90 & 42.42 & 74.00 & 84.35 & 5.59 \\
        LF-Att \cite{das2017visual}        & 49.76 & 57.07 & 42.08 & 74.82 & 85.05 & 5.41 \\
        FGA \cite{Schwartz2019FactorGA}    & 52.10 & 63.70 & 49.58 & \textbf{80.97} & 88.55 & 4.51 \\
        GNN \cite{Zheng2019ReasoningVD}    & 52.82 & 61.37 & 47.33 & 77.98 & 87.83 & 4.57 \\
        CorefNMN \cite{kottur2018visual}   & 54.70 & 61.50 & 47.55 & 78.10 & 88.80 & 4.40 \\
        RvA \cite{Niu_2019_CVPR}           & 55.59 & 63.03 & 49.03 & 80.40 & 89.83 & 4.18 \\
        Synergistic \cite{Guo_2019_CVPR}   & 57.32 & 62.20 & 47.90 & 80.43 & 89.95 & 4.17 \\
        DAN  \cite{Kang2019DualAN}         & 57.59 & 63.20 & 49.63 & 79.75 & 89.35 & 4.30 \\
        \midrule
         \textbf{LTMI}$^\star$                       & 59.03 & \textbf{64.08} & \textbf{50.20} & 80.68 & \textbf{90.35} & \textbf{4.05} \\
        \textbf{LTMI}                               & \textbf{60.92} & 60.65 & 47.00 & 77.03 & 87.75 & 4.90  \\
      \bottomrule
    \end{tabular}
    }
& &
    \raisebox{0ex - \height}{
    \begin{tabular}{rcccccc}
    \multicolumn{7}{c}{\large \centering b) Performance of ensemble models} \\
    \toprule 
    Model                                         & \textbf{NDCG} $\uparrow$  & MRR $\uparrow$  & R@1 $\uparrow$   & R@5 $\uparrow$   & R@10 $\uparrow$   & Mean $\downarrow$ \\
    \midrule
    FGA \cite{Schwartz2019FactorGA}   & 52.10 & 67.30 & 53.40 & 85.28 & 92.70 & 3.54 \\
    Synergistic \cite{Guo_2019_CVPR}         & 57.88 & 63.42 & 49.30 & 80.77 & 90.68 & 3.97 \\
    DAN  \cite{Kang2019DualAN}        & 59.36 & 64.92 & 51.28 & 81.60 & 90.88 & 3.92 \\
    ReDAN \cite{Gan2019MultistepRV}   & 64.47 & 53.73 & 42.45 & 64.68 & 75.68 & 6.63 \\
    \textbf{LTMI}                          & 66.53 & 63.19 & 49.18 & 80.45 & 89.75 & 4.14 \\ 
    \midrule
    P1\_P2\cite{qi2020two}$^{\dagger}$ & 74.91          &	49.13  & 36.68 & 62.98 &    78.55 &    7.03 \\
    VD-BERT\cite{wang2020vd}$^\dagger$  & \textbf{75.13} & 50.00 &	38.28 & 60.93 &	    77.28  &	6.90 \\
    \textbf{LTMI}$^\dagger$              & 74.88 & \textbf{52.14} & \textbf{38.93} & \textbf{66.60} & \textbf{80.65} & \textbf{6.53} \\
    \bottomrule
    
    \\
   \multicolumn{7}{c}{\large \centering c) Num. of attention parameters and the metrics scores} \\
    \midrule
    \multicolumn{2}{r}{Model} & \multicolumn{2}{r}{\# params} & &\MRR & \textbf{NDCG}$\uparrow$ \\
      \midrule 
      \multicolumn{2}{r}{DAN   \cite{Kang2019DualAN}}           && \multicolumn{2}{l}{12.6M} & 63.20  & 57.59 \\
      \multicolumn{2}{r}{RvA   \cite{Niu_2019_CVPR}}            && \multicolumn{2}{l}{11.9M} & 63.03  & 55.59 \\
      \multicolumn{2}{r}{Naive Transformer}                     && \multicolumn{2}{l}{56.8M} & 62.09  & 55.10 \\
      \midrule
      \multicolumn{2}{r}{\textbf{LTMI}* (MRR-based)}                     && \multicolumn{2}{l}{4.8M} &  \textbf{64.08}  & 59.92 \\
      \multicolumn{2}{r}{\textbf{LTMI} (Q, V)}                           && \multicolumn{2}{l}{4.8M} &  60.65  & 60.92  \\
      \multicolumn{2}{r}{\textbf{LTMI} (Q, V, R)}                        && \multicolumn{2}{l}{4.8M} &  60.76  & \textbf{61.12}  \\
      \bottomrule
    
    \end{tabular}
    }
\end{tabular}
\label{tab:result_visdial_1.0}
}
\end{center}
\vspace*{-5mm}
\end{table}
Many previous studies report the performance of an ensemble of multiple models. To make a comparison, we create an ensemble of 16 models with some differences, from initialization with different random seeds to whether to use sharing weights across attention blocks or not, the number of attention blocks (i.e. $L$ = 2, 3), and the number of objects in the image (i.e. $K$ = 50, 100).
Aiming at achieving the best performance, we also enrich the image features by incorporating the class label and attributes of each object in an image, which are also obtained from the pretrained Faster-RCNN model. Details are given in the supplementary material. We take the average of the outputs (probability distributions) from the discriminative decoders of these models to rank the candidate answers. Furthermore, we also test fine-tuning each model with its discriminative decoder on the available dense annotations from the train v1.0 and val v1.0, where the cross-entropy loss with soft labels (i.e. relevance scores) is minimized for two epochs. Table \ref{tab:result_visdial_1.0}(b) shows the results. It is observed that our ensemble model (w/o the fine-tuning) achieves the best $\textrm{NDCG}=66.53$ in all the ensemble models.

With optional fine-tuning, our ensemble model further gains a large improvement in NDCG, resulting in the third place in the leaderboard. 
The gap in NDCG to the first place (VD-BERT) is only 0.25pp, while our model yields performance that is better in all the other metrics, i.e, by 2.14pp, 5.67pp, and 3.37pp in MRR, R@5, and R@10, respectively, and 5.36\% reduction in Mean. 

Table \ref{tab:result_visdial_1.0}(c) shows the number of parameters of the multi-modal attention mechanism employed in the recent methods along with their NDCG scores on the VisDial v1.0 test-standard set. We exclude the parameters of the networks computing the input utilities and the decoders, as they are basically shared among these methods. `Naive Transformer' consists of two stacks of transformer blocks with simple extension to three utilities as mentioned in Sec.~\ref{introduction}. The efficiency of our models can be observed. 
Note also that the gap between (Q, V) and (Q, V, R) is small, contrary to the argument in \cite{qi2020two}.
\begin{table}[t]
    \caption{ Ablation study on the components of our method on the val v1.0 split of VisDial dataset. $\uparrow$ indicates the higher the better.}
    \vskip -0.1in
    \resizebox{1.0\columnwidth}{!}{
    \begin{tabular}{ccc}
        \raisebox{0ex - \height}{
            \begin{tabular}{l|c|ccc}
                \multicolumn{5}{c}{\normalsize \centering (a)  
                }\\
                \toprule
                Component & Details & A-NDCG $\uparrow$ & D-NDCG  $\uparrow$ & G-NDCG $\uparrow$  \\
                \midrule
                Number of           & 1     & 65.37  & 62.06 & 62.95 \\
                attention blocks    & 2     & \textbf{65.75}  & \textbf{62.72} & \textbf{63.58} \\
                                    & 3     & 65.42  & 62.48 & 63.22   \\
                \midrule
                Self-Attention      &  No   & 65.38  & 61.76 & 63.31           \\
                                    &  Yes  & \textbf{65.75}  & \textbf{62.72} & \textbf{63.58} \\
                \midrule
                Attended features   &  Add     & 64.12  & 60.28 & 61.49           \\
                aggregation         &  Concat  & \textbf{65.75}  & \textbf{62.72} & \textbf{63.58} \\
                
                \midrule 
                Shared Attention    & No    & \textbf{65.75}  & \textbf{62.72} & \textbf{63.58} \\ 
                       weights      & Yes   & 65.57  & 62.50 & 63.24  \\          
                    
            \bottomrule
            \end{tabular}
        }
    & &
        \raisebox{0ex - \height}{
            \begin{tabular}{l|c|ccc}
            \multicolumn{5}{c}{\normalsize \centering (b)}\\
            \toprule
            Component & Details & A-NDCG $\uparrow$ & D-NDCG  $\uparrow$ & G-NDCG $\uparrow$  \\

            \midrule
            Context feature     &  [Q]       & 65.12  & 61.50 & 63.19           \\
            aggregation         &  [Q, V]    & \textbf{65.75}  & \textbf{62.72} & \textbf{63.58} \\
                                &  [Q, V, R] & 65.53  & 62.37 & 63.38            \\
            \midrule
            Decoder Type        & Gen     & -  & - & 62.35 \\
                                & Disc    & -  & 61.80 & - \\
                                & Both    & \textbf{65.75}  & \textbf{62.72} & \textbf{63.58} \\
            \midrule
            The number of           & 36        & 65.25  & 62.40 & 63.08  \\
            objects in an image     & 50        & 65.24  & 62.29 & 63.12  \\
                                    & 100  & \textbf{65.75}  & \textbf{62.72} & \textbf{63.58} \\         
            \midrule 
            Positional and         & No    & 65.18  & 61.84 & 62.96           \\
            spatial embeddings     & Yes   & \textbf{65.75}  & \textbf{62.72} & \textbf{63.58} \\
            
            \bottomrule 
            \end{tabular}
        }
    \end{tabular}
    \label{tab:visdial_0.9_ablative}
    }
    \vskip -0.1in
    \end{table}
\subsection{Ablation Study} \label{sec:abalation_study}
To evaluate the effect of each of the components of our method, we perform the ablation study on the val v1.0 split of VisDial dataset. We evaluate here the accuracy of the discriminative decoder and the generative decoder separately. We denote the former by D-NDCG and the latter by G-NDCG, and the accuracy of their averaged model by A-NDCG (i.e., averaging the probability distributions over the candidate answers obtained by the discriminative and generative decoders). The results are shown in Table \ref{tab:visdial_0.9_ablative}(a-b).

The first block of Table \ref{tab:visdial_0.9_ablative}(a) shows the effect of the number of stacks of the proposed attention blocks. We observe that the use of two to three stacks achieves good performance on all three measures. More stacks did not bring further improvement, and thus are omitted in the table. 

The second block of Table \ref{tab:visdial_0.9_ablative}(a) shows the effect of self-attention, which computes the interaction within a utility, i.e., ${\bar{\cal A}}_X(X)$. 
We examine this because it can be removed from the attention computation. It is seen that self-attention does contribute to good performance. The third block shows the effects of how to aggregate the attended features.
It is seen that their concatenation yields better performance than their simple addition. The fourth block shows the impact of sharing the weights across the stacks of the attention blocks. If the weights can be shared as in \cite{lan2019albert}, it contributes a further decrease in the number of parameters. We observe that the performance does drop if weight sharing is employed, but the drop is not very large.

The first block of Table \ref{tab:visdial_0.9_ablative}(b) shows the effect of how to aggregate the context features $c_V$, $c_Q$, and $c_R$ in the decoder(s), which are obtained from the outputs of our encoder. As mentioned above, the context vector $c_R$ of the dialog history does not contribute to the performance. However, the context vector $c_v$ of the image is important for achieving the best performance. The second block of Table \ref{tab:visdial_0.9_ablative}(b) shows the effects of simultaneously training the both decoders (with the entire model). It is seen that this contributes greatly to the performance; this indicates the synergy of learning two tasks while sharing the encoder, resulting better generalization as compared with those trained with a single decoder. 

We have also confirmed that the use of fewer objects leads to worse results. Besides, the positional embedding for representing the question and history utilities as well as the spatial embedding (i.e., the bounding box geometry of objects) for image utility representation have a certain amount of contribution. 
  
\subsection{Visualization of Generated Attention}\label{sec:visualization}
\begin{figure}[t]
\centering
\includegraphics[width=1.0\textwidth]{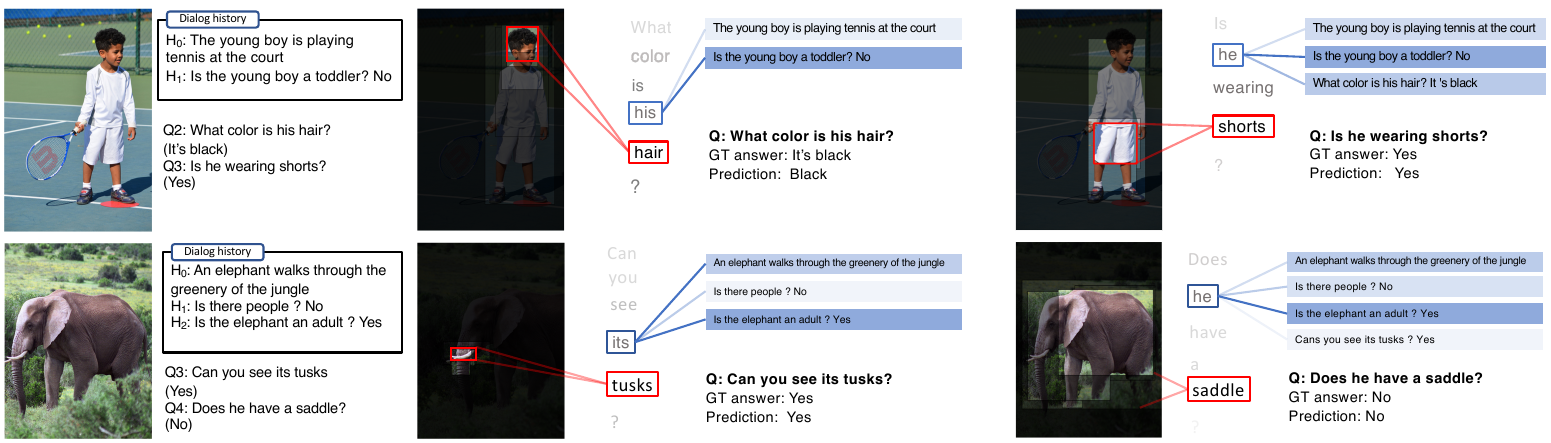}
\caption{\color{black}Examples of visualization for the attention weights
generated in our model at two Q\&A rounds on two images. See Sec.~\ref{sec:visualization} for details.
}
\label{fig:qualitative_results}
\vskip -0.1in
\end{figure}
Figure \ref{fig:qualitative_results} shows attention weights generated in our model on two rounds of Q\&A on two images. We show here two types of attention. One is the self-attention weights used to compute the context vectors $c_V$ and $c_Q$. For $c_V$, the attention weights $a_{V}$ are generated over image regions (i.e., bounding boxes), as in Eq.(\ref{eqn:a_v_weights}). Similarly, for $c_Q$, the attention weights are generated over question words. 
These two sets of attention weights are displayed by brightness of the image bounding-boxes and darkness of question words, respectively, in the center and the rightmost columns. It can be observed from these that the relevant regions and words are properly highlighted at each Q\&A round. 

The other attention we visualize is the source-to-target attention computed inside the proposed block. We choose here the image-to-question attention $\bar{\cal A}_V(Q)$ and the history-to-question attention $\bar{\cal A}_R(Q)$. For each, we compute the average of the attention weights over all the heads computed inside the block belonging to the upper stack. 
In Fig.~\ref{fig:qualitative_results}, the former is displayed by the red boxes connected between an image region and a question word; only the region with the largest weight is shown for the target word; the word with the largest self-attention weight is chosen for the target. The history-to-question attention is displayed by the Q\&As highlighted in blue color connected to a selected question word that is semantically ambiguous, e.g., {\em `its'}, {\em `he'}, and {\em `his'}. It is seen that the model performs proper visual grounding for the important words, {\em`hair'}, {\em`shorts'}, and {\em'tusks'}. It is also observed that the model properly resolves the co-reference for the words, {\em `he'} and {\em `its'}. 

\section{Summary and Conclusion}

In this paper, we have proposed LTMI (Light-weight Transformer for Many Inputs)
that can deal with all the interactions between multiple input utilities in an efficient way. As compared with other methods, the proposed architecture is much simpler in terms of the number of parameters as well as the way of handling inputs (i.e., their equal treatment), and nevertheless surpasses the previous methods in accuracy; it achieves the new state-of-the-art results on the VisDial datasets, e.g., high NDCG scores on the VisDial v1.0 dataset. Thus, we believe our method can be used as a simple yet strong baseline.

\bigskip
\noindent
{\bf Acknowledgments}~
This work was partly supported by JSPS KAKENHI Grant Number JP15H05919 and JP19H01110..

\bibliographystyle{splncs04}
\bibliography{egbib}

\newpage
\appendix
{%
 \centering
 \Large \textbf{Supplementary Material}\\[1.5em]
}

\section{Representations of Utilities}
\subsection{Image Utility}

The image utility is represented by the standard method employed in many recent studies. It is based on the bottom-up mechanism \cite{anderson2018bottom}, which extracts region-level image features using the Faster-RCNN pre-trained on the Visual Genome dataset \cite{krishna2017visual}. For each input image, we select the top $K$ objects, and represent each of them by a visual feature $v^r_i\in\mathbb{R}^{2048}$ and a bounding box expressed by $(x_{i,1},x_{i,2})$ and $(x_{i,3},x_{i,4})$ (the coordinates of the upper-left and lower-right corners.) 

The feature vector $v^r_i$ is then converted into another vector $v^f_i\in\mathbb{R}^d$ as follows. We introduce the following notation to express a single FC layer with ReLU, to which dropout regularization is applied:
\begin{equation}
    \underset{k \rightarrow d}{\mathrm{MLP}}(x) \equiv \mathrm{Dropout}(\mathrm{ReLU}(W^{\top}x + b)),
\end{equation}
where $x \in \mathbb{R}^{k}$ is the input and $W \in \mathbb{R}^{k \times d}$ and $b\in \mathbb{R}^d$ are the weights and biases. Then, $v^f_i$ is obtained by \begin{equation}
    v^f_i = \mathrm{LayerNorm}(\mathop{\mathrm{MLP}}_{2048 \rightarrow d}(v^r_i)), 
\end{equation}
where LayerNorm is the layer normalization \cite{ba2016layer} applied to the output. 

The bounding box geometry is converted into $v^b_i\in \mathbb{R}^d$ in the following way. First, the image is resized to $600\times 600$ pixels and the bounding box geometry is transformed accordingly. Then, representing each of the four coordinates by a one-hot vector of size 600, we convert them into the embedding vectors  $\hat{x}_{i,1},\ldots,\hat{x}_{i,4}(\in\mathbb{R}^{d})$ using four different embedding layers. Then, we obtain $v^b_i$ as below
\begin{equation}
    v^b_i = \sum_{j=1}^{4}\mathrm{LayerNorm}(
    \mathop{\mathrm{MLP}}_{d\rightarrow d}(\hat{x}_{i,j})).
\end{equation}

Finally, $v^f_i$ encoding the visual feature and $v^b_i$ encoding the spatial feature are aggregated by adding and normalizing as
\begin{equation}\label{eq:agg}
    v_i = \mathrm{LayerNorm}(v^f_i + v^b_i). 
\end{equation}
The resulting $v_i$'s for the $K$ objects ($i=1,\ldots,K$) comprise a matrix $V = [v_1, v_2, \cdots, v_K]^\top \in \mathbb{R}^{K \times d}$, which gives the representation of the visual utility.

\paragraph{Optional Image Feature Enrichment.} 
In the experiment of comparing ensembles on the test split of Visdial v1.0, we enrich the image features for further improvement. 
To be specific, for each object, we also obtain a class label with highest probability (e.g. {\em`cat', `hair', and `car'}) and the top 20 attributes for each class label (e.g., `curly', `blond', `long', and so on, for the label `hair'). These can be extracted from the Faster-RCNN along with the above CNN features and bounding box geometry. We incorporate these into the image utility representation in the following way.

The class label for the $i$-th object is first encoded into an embedding vector $e_i^c \in \mathbb{R}^{300}$ using the same embedding layer as the question. Then, we convert $e_i^c$ into a $d$-dimensional vector $v_i^c$ by
\begin{equation}
    v_i^c = \mathrm{LayerNorm}(\mathop{\mathrm{MLP}}_{300\rightarrow d}(e_i^c)).
\end{equation}
Similarly, for the top 20 attributes of each object $i$, we encode them into embedding vectors of size $300$, i.e. $e_{i,1}^a, \ldots, e_{i,20}^a$, and then convert them further into $v^a_i \in \mathbb{R}^{d}$ as 
\begin{equation}
    v^a_i = \sum_{j=1}^{20}\mathrm{LayerNorm}(
    \mathop{\mathrm{MLP}}_{300\rightarrow d}(e_{i,j}^a)w_{i,j}^a,
\end{equation}
where $w_{i,j}^a$ is the confidence score extracted from the Faster-RCNN for attribute $j$ of the $i$-th object. Then, the visual feature $v_i^f$, the spatial feature $v_i^b$, the class feature $v_i^c$, and the attribute feature $v_i^a$ are aggregated by their addition followed by normalization as
\begin{equation}
        v_i = \mathrm{LayerNorm}(v^f_i + v^b_i + v_i^c + v_i^a). 
\end{equation}
We then use these vectors to form the matrix $V$ instead of Eq.(\ref{eq:agg}).

\subsection{Question Utility} \label{ques_utility}

The question utility is also obtained by the standard method but with one exception, the employment of positional embedding used in NLP studies. Note that we examine its effects in an ablation test shown in the main paper. A given question sentence is first fit into a sequence of $N$ words; zero-padding is applied if necessary.
Each word $w_i$ ($i=1,\ldots,N)$ is embedded into a vector $e_i$ of a fixed size using an embedding layer initialized with pretrained GloVe vectors \cite{pennington2014glove}. They are then inputted into two-layer Bi-LSTM, obtaining two d-dimensional vectors $\overrightarrow{h_{i}}$ and $\overleftarrow{h_{i}}$ as their higher-layer hidden state: 
\begin{equation}
      \begin{aligned}
    \overrightarrow{h_{i}} & =  \mathrm{LSTM}(e_i, \overrightarrow{h_{i-1}}),\\
        \overleftarrow{h_{i}} & =  \mathrm{LSTM}(e_i, \overleftarrow{h_{i+1}}).
    \label{eq:ques_lstm}
  \end{aligned}
\end{equation}
Their concatenation, $h_{i} = [ \overrightarrow{h_{i}}^\top, \overleftarrow{h_{i}}^\top ]^\top$, is then projected back to a $d$-dimensional space using a linear transformation, yielding a vector $q^f_i$. Positional embedding $q^p_i$ from the paper \cite{vaswani2017attention} is added to get the final representation $q_i \in \mathbb{R}^{d}$ of $w_i$ as 
\begin{equation}
    q_{i} = \mathrm{LayerNorm}(q^f_i + q^p_i).
    \label{eq:ques_layernorm}
\end{equation}
The representation of the question utility is given as $Q = [q_1,\ldots,q_N]^\top \in \mathbb{R}^{N \times d}$.

\subsection{Dialog History Utility} \label{hist_utility}

In this study, we choose to represent the dialog history as a single utility. Each of its entities represents the question-answer pair at one round. As with previous studies, the caption is treated as the first round of $2N$-word which is padded or truncated if necessary. For each round $t > 1$, the word sequences of the question and the answer at the round is concatenated into $2N$-word sequence with zero padding if necessary.
As with the question utility, after embedding each word into a GloVe vector, the resulting sequence of $2N$ embedded vectors is inputted to two-layer Bi-LSTM, from which only their last (higher-layer) hidden states are extracted to construct $2d$-dimensional vector $[\overrightarrow{h_{0}}^\top, \overleftarrow{h_{2N}}^\top]^\top$. We then project it with a linear transform to a $d$-dimensional space, yielding $r^f_t\in \mathbb{R}^d$. For the linear projection, we use different learnable weights from the question utility. As in Eq.(\ref{eq:ques_layernorm}), we add positional embedding, which represents the order of rounds, and then apply layer normalization, yielding a feature vector of the round $t$ question-answer pair. The history utility is then given by $R = [r_1,\ldots,r_T]^\top \in \mathbb{R}^{T \times d}$.

\section{Design of Decoders}

\subsection{Discriminative Decoder}
A discriminative decoder outputs the likelihood score for each of 100 candidate answers for the current question at round $T$ in the following way. We use a similar architecture to the one used to extract question features in Sec.~\ref{ques_utility} to convert each candidate answer (indexed by $i(=1,\ldots,100)$) to a feature vector $a_i \in \mathbb{R}^d$. Specifically, it is two-layer Bi-LSTM receiving a candidate answer at its input, on top of which there is a linear projection layer followed by layer normalization. Using the resulting vectors, the score $p_i$ for $i$-th candidate answer is computed by 
\begin{equation}
p_i=\mathrm{logsoftmax}_i(a_1^\top c,\ldots,a_{100}^\top c). 
\end{equation}
In the test phase, we sort the candidate answers using these scores. In the training phase, the cross-entropy loss $\mathcal{L}_D$ between $p=[p_1,\ldots,p_{100}]^\top$ and the ground truth label encoded by a one-hot vector $y$ is minimized:
\begin{equation}\label{eqn:disc_loss}
    \mathcal{L}_D = -\sum_{i=1}^{100}y_i p_i.
\end{equation}
When relevance scores $s=[s_1,\ldots,s_{100}]^\top$ over the answer candidates are available (called dense annotation in the VisDial dataset) rather than a single ground truth answer, we can use them by setting $y_i=s_i$ for all $i$'s and minimize the above loss. We employ dropout with rate of 0.1 for the LSTM.

\subsection{Generative Decoder} \label{sec:gen_decoder}

Following \cite{das2017visual}, we also consider a generative decoder to score the candidate answers using the log-likelihood scores. The generative decoder consists of a two-layer LSTM  to generate an answer using the context vector $c$ as the initial hidden state. In the training phase, we predict the next token based on the current token from the ground truth answer. In details, we first append the special token ``SOS" at the beginning of the ground truth answer, then embedding all the sentence into the embedding vectors $a_{gt} = [w_0, w_1, \ldots, w_N]$ where $w_0$ is the embedding vector of ``SOS" token. The hidden state $h_n \in \mathbb{R}^{d}$ at the $n$-th timestep (extracted from the higher-layer LSTM) is computed given $w_{n-1}$ and $h_{n-1}$ as follows:
\begin{equation}
    h_n = \textrm{LSTM}(w_{n-1}, h_{n-1}),   
\end{equation}
where $h_0$ is initialized by $c$. Thus, we compute $p_n$, the log-likelihood of $n$-th word as 
\begin{equation} \label{eq:p_n}
    p_n = \mathrm{logsoftmax}_j(W_n^\top h_n  + b_n),
\end{equation}
where $W_n \in \mathbb{R}^{d \times |V|}$ and $p_n \in \mathbb{R}^{|V|}$, where $|V|$ is the vocabulary size; and $j$ is the index of $n$-th word in the vocabulary.

In the training phase, we minimize $\mathcal{L}_G$, the summation of the negative log-likelihood defined by
\begin{equation}
    \mathcal{L}_G = -\sum_{n=1}^{N}p_n.
\end{equation}
In the validation and test phase, for each candidate answer $A_{T,i}$, we compute $s_i = \sum_{n=1}^{N}p_n^{(A_{T,i})}$ where $p_n^{(A_{T,i})}$ is the log-likelihood of the $n$-th word in the candidate answer $A_{T,i}$ which is computed similarly as in Eq.(\ref{eq:p_n}). Then, the rankings of the candidate answers are derived as $\mathrm{softmax}_{i}(s_1, \ldots, s_{100})$. We employ dropout with rate of 0.1 for the LSTM.

\section{Implementation Details}

When computing $\bar{\cal A}_{Y}(X)$, we perform the following form of computation
\[
{\cal A}(Q,K,V)=\mathrm{softmax}\left(
\frac{Q K^\top}{\sqrt{d}}
\right)V,
\]
where we compute a matrix product $Q K^\top$ as above. In the computation of $\bar{\cal A}_{X}(Y)$, we need another matrix product, but it is merely the transposed matrix $K Q^\top$ due to the symmetry between $X$ and $Y$. 
For the computational efficiency, we perform computation of $\bar{\cal A}_{Y}(X)$ and $\bar{\cal A}_{X}(Y)$ simultaneously; see  $\texttt{MultiHeadAttention}(X, Y)$ in our code. 
Further, following \cite{nguyen2018improved}, we also pad $X$ and $Y$ with two $d$-dimensional vectors that are randomly initialized with He normal initialization.  This implements ``no-where-to-attend'' features in the computation of $\bar{\cal A}_{Y}(X)$ and $\bar{\cal A}_{X}(Y)$.

\begin{table}[h]
    \centering
    \caption{Hyperparamters used in the training procedure.}
    \begin{tabular}{ll}
    \toprule
    Hyperparameter          &  Value \\
    \midrule
    Warm-up learning rate   &  $1\mathrm{e}{-5}$ \\
    Warm-up factor & 0.2 \\
    Initial learning rate after the 1st epoch & $1\mathrm{e}{-3}$ \\
    $\beta_1$ in Adam & 0.9 \\
    $\beta_2$ in Adam & 0.997 \\
    $\epsilon$ in Adam & $1\mathrm{e}{-9}$  \\
    Weight decay & $1\mathrm{e}{-5}$  \\
    Number of workers & 8 \\
    Batch size & 32 \\
    \bottomrule
    \end{tabular}
    \label{tab:hyperparams}
\end{table}

Table \ref{tab:hyperparams} shows the hyperparameters used in our experiments, which are selected following the previous studies. We perform all the experiments on a GPU server that has four Tesla V100-SXM2 of 16GB memory with CUDA version 10.0 and Driver version 410.104. It has Intel(R) Xeon(R) Gold 6148 CPU @ 2.40GHz of 80 cores with the RAM of 376GB memory. We use Pytorch version 1.2 \cite{paszke2017automatic} as the deep learning framework. 

\section{Analysis on Visdial v1.0 Validation Split}

\subsection{
Analyzing Inconsistency between NDCG and Other Metrics}
\begin{figure*}[h]
\centering
\centerline{
\includegraphics[width=1.25\textwidth]{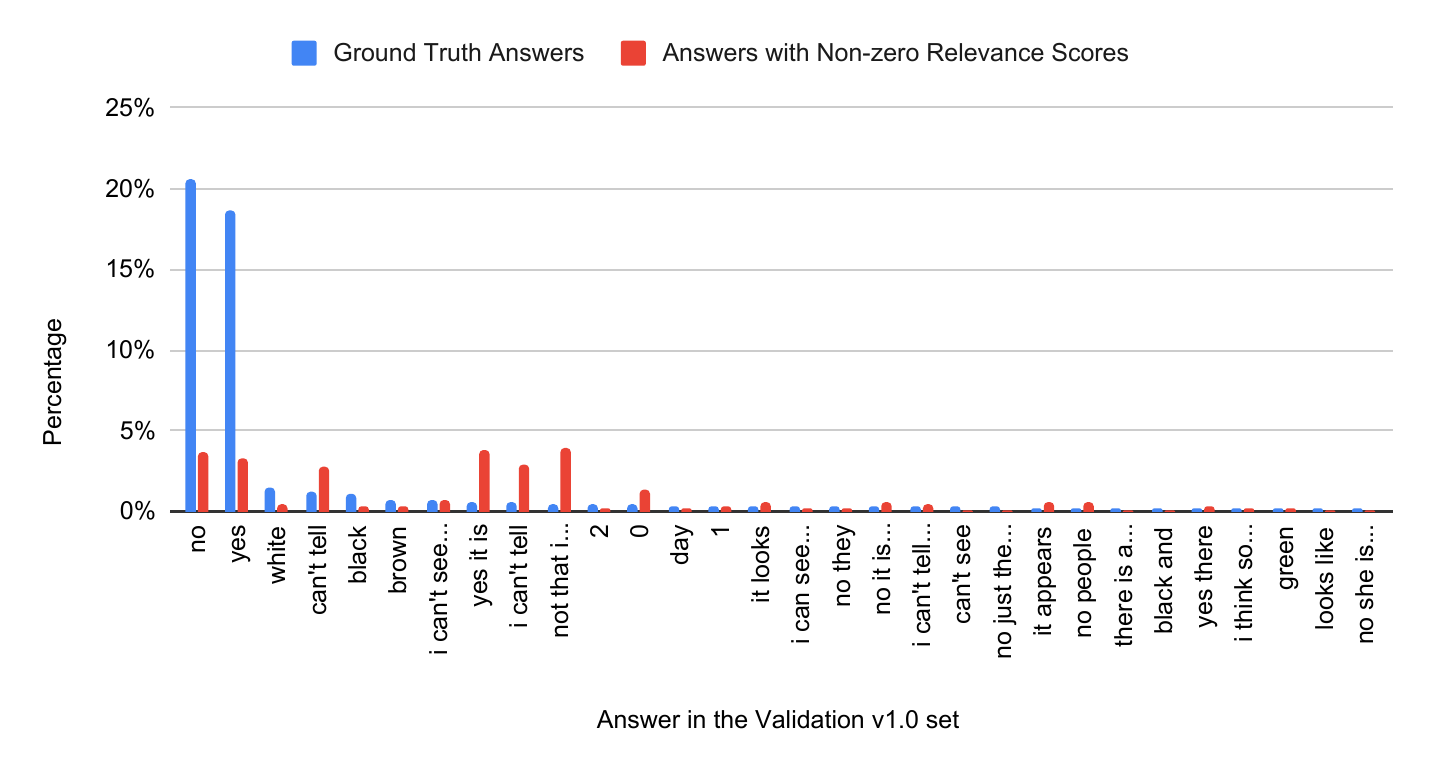}
}
\caption{The distribution of the 30 most popular answers in the Validation v1.0 set.}
\label{fig:freq_chart}
\end{figure*}

As mentioned in the main paper, we observed the inconsistency in the performance of models evaluated by NDCG and other metrics, such as MRR. The same is also reported in recent studies. We show an analysis on this. 

We first recap how the Visdial v1.0 dataset was collected \cite{das2017visual}. A live chat between two workers, i.e., a questioner and an answerer, was conducted on Amazon Mechanical Turk (AMT). For an image provided with a caption, the questioner raised a question 
based on the caption without seeing the image. The answerer responded to the question by looking at the image, which are used as ground truth answers. 

To cope with the difficulty of evaluating answers generated by a model in the form of free texts, 
Das et al. \cite{das2017visual} proposed a method that discriminatively evaluates the performance of visual dialog systems by using a set of 100 candidate answers, to each of which a relevance score is given. It makes a system under evaluation return the rankings of all the candidate answers and then calculates  {
the scores of metrics, e.g. NDCG, MRR, etc.}
based on the returned rankings. 
To create a set of 100 candidate answers for each question, they
collected from all the answers given by the answerers, the \textbf{plausible} answers of the 50 most similar questions to the ground truth answer including itself, the 30 most \textbf{popular} answers, and 20 
\textbf{random} answers. Each of these candidate answers was then 
given a relevance score with a consensus of several AMT workers.

Now we make a few observations on the dataset. 
First, the ground truth answers provided by the answerers are not always high-quality. As shown in Table \ref{tab:gt_answers}, 
33.6\% of the ground truth answers have relevance scores lower than 0.5. 
Assuming the AMT workers giving the relevance scores to be accurate, this implies some of the ``ground truth'' answers provided by the answerers are simply wrong.  
Second, the answerers tend to  more frequently use short, general answers, such as `\textit{no}' and `\textit{yes}'. 
This is illustrated in Fig.~\ref{fig:freq_chart} that shows
the frequencies of the most popular ones in the ground truth answers and also those having non-zero relevance scores.
It is clearly seen that the short and less informative answers (i.e., `\textit{no}' and `\textit{yes}') are less frequently considered to be relevant. 

Recall that NDCG metric is measured based on the rankings of all the candidate answers, whereas MRR and other metrics are based on the ranking of the ground truth answers. Based on the above observations, we can say that the NDCG is more appropriate as an evaluation metric, following the other recent studies. This claim is also supported by Table \ref{tab:gt_answers}, the results of the experiments examining how evaluated performances vary depending on when to stop the training of the proposed model. It is seen from the table that the model at epoch 5, which is noted as `MRR-favored' as it corresponds to early stopping based on validation on MRR, yields high MRR and low NDCG scores over all questions. 
It tends to give higher scores on the safe and popular answers that appear more frequently in the ground truth answers. When we continue to train the model until 12 epochs, it (noted as NDCG-favored) generates better rankings for all possible answers rather than only the ground truth answers, yielding large improvements in NDCG scores. However, it yields lower MRR scores, since the model does not give high ranks to some of the ``ground truth'' answers; they are indeed very likely to be bad answers. It is also seen from the table that the both models yield better scores on the both MRR and NDCG metrics for the questions having the ground truth answers with high relevance scores.

\begin{table}[t]
\caption{The performance of the training strategies, i.e. based on the MRR or NDCG early stopping, categorized by questions of corresponding relevance score of ground truth answers.}
\begin{center}
    \begin{tabular}{
        >{\centering}p{0.15\textwidth} 
        >{\centering}p{0.15\textwidth} 
        >{\centering}p{0.12\textwidth} 
        >{\centering}p{0.12\textwidth} 
        >{\centering}p{0.01\textwidth} 
        >{\centering}p{0.12\textwidth} 
        >{\centering\arraybackslash}p{0.12\textwidth} 
        }
        \toprule
        \multirow{2}{*}{Rel Score}  & \multirow{2}{*}{Percentage} & \multicolumn{2}{c}{MRR-favored} & & \multicolumn{2}{c}{NDCG-favored} \\
        \cmidrule{3-4}
        \cmidrule{6-7}
    
                     && MRR&NDCG && MRR&NDCG                \\
        \midrule
        0.0       & 9.0\%      & \textbf{58.26}      & 44.06     && 56.12      & \textbf{48.00}      \\
        0.2       & 11.0\%     & \textbf{58.26}      & 44.06     && 57.70      & \textbf{54.46}      \\
        0.4       & 13.6\%     & \textbf{61.07}      & 56.94     && 60.69      & \textbf{58.94}      \\
        0.6       & 16.0\%     & \textbf{65.38}      & 59.40     && 62.35      & \textbf{62.37}      \\
        0.8       & 19.2\%     & \textbf{67.34}      & 62.90     && 64.45      & \textbf{66.79}      \\
        1.0       & 31.2\%     & \textbf{67.48}      & 65.13     && 65.37      & \textbf{69.21}      \\
        \bottomrule
    \end{tabular}
    \end{center}
\label{tab:gt_answers}
\end{table}

\subsection{Question-Type Analysis}
\begin{table}[t]
\caption{The performance comparison of discriminative and generative decoders evaluated on question types evaluated on the NDCG metric.}
\begin{adjustbox}{width=0.6\columnwidth,center}
\begin{tabular}{llllll}
\toprule
\multicolumn{2}{l}{Question Type}                                  & Yes/No         & Number         & Color          & Others         \\
\midrule
\multicolumn{2}{l}{Percentage}                                     & 75\%           & 3\%            & 11\%           & 11\%           \\
\midrule
Decoder                         & Model                            &                &                &                &                \\
\midrule
\multirow{2}{*}{Generative}     & ReDAN \cite{Gan2019MultistepRV}  & 63.49          & 41.09          & 52.16          & 51.45          \\
                                & Ours                             & \textbf{66.24} & \textbf{46.35} & \textbf{55.77} & \textbf{57.25} \\
\midrule
\multirow{2}{*}{Discriminative} & ReDAN \cite{Gan2019MultistepRV}  & 60.89          & 44.47          & 58.13          & 52.68          \\
                                & Ours                             & \textbf{64.08} & \textbf{49.86} & \textbf{60.95} & \textbf{58.16}
    \\
\bottomrule
\end{tabular}
\end{adjustbox}
\label{tab:question_type}
\end{table}

\begin{table}[ht!]
    \caption{Retrieval performance of compared methods and ours on the val v0.9 split reported with a single model.}
    \begin{center}
    \begin{adjustbox}{width=0.6\columnwidth,center}
        \begin{tabular}{rccccc}
        \toprule
        Model & MRR $\uparrow$  & R@1 $\uparrow$    & R@5 $\uparrow$  & R@10 $\uparrow$  & Mean $\downarrow$ \\
        \midrule
        SAN \cite{yang2016stacked}          & 57.64 & 43.44 & 74.26 & 83.72 & 5.88 \\
        LF \cite{das2017visual}             & 58.07 & 43.82 & 74.68 & 84.07 & 5.78 \\
        HRE \cite{das2017visual}            & 58.46 & 44.67 & 74.5  & 84.22 & 5.72 \\
        HREA \cite{das2017visual}           & 58.68 & 44.82 & 74.81 & 84.36 & 5.66 \\
        MN \cite{das2017visual}             & 59.65 & 45.55 & 76.22 & 85.37 & 5.46 \\
        NMN \cite{hu2017learning}           & 61.60 & 48.28 & 77.54 & 86.75 & 4.98 \\
        HCIAE \cite{lu2017best}             & 62.22 & 48.48 & 78.75 & 87.59 & 4.81 \\
        AMEM  \cite{seo2017visual}          & 62.27 & 48.53 & 78.66 & 87.43 & 4.86 \\
        SF \cite{jain2018two}               & 62.42 & 48.55 & 78.75 & 87.75 & 4.47 \\
        GNN \cite{Zheng2019ReasoningVD}     & 62.85 & 48.95 & 79.65 & 88.36 & 4.57 \\
        CoAtt \cite{wu2018you}              & 63.98 & 50.29 & 80.71 & 88.81 & 4.47 \\
        CoefNMN \cite{kottur2018visual}     & 64.10 & 50.92 & 80.18 & 88.81 & 4.45 \\
        FGA \cite{Schwartz2019FactorGA}     & 65.25 & 51.43 & 82.08 & 89.56 & 4.35 \\
        RvA \cite{Niu_2019_CVPR}            & 66.34 & 52.71 & 82.97 & 90.73 & 3.93 \\
        DAN \cite{Kang2019DualAN}           & 66.38 & 53.33 & 82.42 & 90.38 & 4.04 \\   
        \midrule
        Ours    & \textbf{67.94} & \textbf{55.05}    & \textbf{83.98} & \textbf{91.58}  & \textbf{3.69} \\
        \bottomrule
        \end{tabular}
    \end{adjustbox}
    \end{center}
    \label{tab:result_v0.9}
\end{table}

Following \cite{Gan2019MultistepRV}, we perform a question-type analysis of the NDCG scores achieved by different decoders from the model mentioned in our main paper. The questions are classified into four categories: Yes/No, Number, Color, and Others. As shown in Table \ref{tab:question_type}, the Yes/No questions account for the majority whereas there is only 3\% of the Number questions. Therefore, the performance on the Yes/No questions translates into the overall performance of any models. Similar to ReDAN \cite{Gan2019MultistepRV}, the performance of ours on the Number questions is the lowest among the other question types, reflecting the hardness of the counting task. 
Another similarity observed on our models and ReDAN is that generative decoders show better performance on the Yes/No questions while discriminative decoders yield higher NDCG scores on the other questions. It is because generative decoders favor the short answers 
that are relevant more often in the Yes/No questions.
It is also seen that our model consistently shows better performance over all question types, i.e. about 3pp on the Yes/No and Color questions, 5pp on the Number questions, and 6pp on the other questions. 

{
\section{Results on the Visdial v0.9 dataset}
Following the previous studies, we report the performance of our method (specifically, the discriminative decoder) on the VisDial v0.9 dataset. The v0.9 dataset consists of the train v0.9 split (82,783 images) and the val v0.9 split (40,504 images).
Note that all the hyperparameter settings are the same as those on the Visdial v1.0 dataset except that we train the model with only five epochs.

Table \ref{tab:result_v0.9} shows the results on the validation set along with performances of other methods. It shows that our model consistently outperforms all the methods across all metrics: MRR, R@1, R@5, R@10 and Mean. 
}

\section{Qualitative Results}
We provide additional examples of the results obtained by our method in Figs.~\ref{fig:s1}-\ref{fig:d5}. They are divided into two groups, results for which the top-1 prediction coincides with the ground truth answer (Figs.~\ref{fig:s1}-\ref{fig:s3}) and those for which they do not coincide (Figs.~\ref{fig:d2}-\ref{fig:d3}). For each result, we show the attention maps created on the input image and question, respectively.

\section{Experiments on AVSD}
{
To test the generality of the proposed method on other tasks as well as its performance on a greater number of utilities, we additionally apply it to the Audio Visual Scene-aware Dialog (AVSD) task \cite{hori2019end}. This task requires a system to generate an answer to a question about events seen in a video given with a previous dialog. AVSD provides more utilities than Visual Dialog, i.e.,  audio features and video features, such as VGG or I3D features (I3D RGB sequence and I3D flow sequence). We build a network by simply replacing the multimodal attention mechanism in the baseline model of \cite{hori2019end} with a simple extension of the proposed attention mechanism. Details are given below.}

\subsection{Network Design}
\begin{table*}[t]
\caption{Comparison of response generation evaluation results with objective measures.}
\begin{adjustbox}{width=1.0\columnwidth,center}
    \begin{tabular}{lc|ccccccc}
    \toprule
    Model                                        & Video Feat. & CIDEr          & BLEU1          & BLEU2          & BLEU3          & BLEU4          & METEOR         & ROUGE\_L       \\
    \midrule
    Baseline \cite{hori2019end} & VGG         & 0.618          & 0.231          & 0.141          & 0.095          & 0.067          & 0.102          & 0.259          \\
    Ours                        & VGG         & \textbf{0.841} & \textbf{0.266} & \textbf{0.172} & \textbf{0.118} & \textbf{0.086} & \textbf{0.117} & \textbf{0.296} \\
    \midrule
    Baseline \cite{hori2019end} & I3D         & 0.727          & 0.256          & 0.161          & 0.109          & 0.078          & 0.113          & 0.277          \\
    Ours                        & I3D         & \textbf{0.851} & \textbf{0.277} & \textbf{0.178} & \textbf{0.122} & \textbf{0.088} & \textbf{0.119} & \textbf{0.302} \\
    \bottomrule
    \end{tabular}
\end{adjustbox}
\label{tab:avsd}
\end{table*}
Following the baselines \cite{hori2019end}, we extract the question utility $Q$ using a two-layer LSTM. We separate the caption from the dialog history and feed it into another two-layer LSTM to obtain the caption utility $C$. Similar to \cite{hori2019end}, the dialog history consisting of previous question-answer pairs is inputted into a hierarchical LSTM network; specifically, we encode each question-answer pair with one LSTM and summarize the obtained encodings with another LSTM, yielding a final vector representation $c_r$. All LSTMs used for language encoding have $d$ units. We convert words into vectors with a shared embedding layer initialized with GLoVe vectors.

The video provides two sources of features, i.e., video features and audio features. We use the audio features extracted from the pretrained VGGish model \cite{hori2019end}, which are fed to a projection layer, providing the audio utility $A$; it is represented as a collection of $d$-dimensional vectors. For video processing, following \cite{hori2019end}, we consider two models with different features: i) VGG features extracted from four uniformly sampled frames in the video, giving the video utility $V$, and ii) I3D features extracted by the I3D network pretrained on an action recognition task, which are forwarded to  projection layers to obtain an I3D-rgb utility and an I3D-flow utility denoted by $V$ and $F$.

To compute the multimodal attention between $U$ utilities, we add a stack of $U$ proposed attention blocks; $U = 4$ for the model (i) and $U=5$ for (ii).
To make the designs of two models (i) and (ii) similar, we use only $A$ utility to attend language utilities; and only $Q$ and $C$ are allowed to attend audio and video utilities. After obtaining the updated representations of all utilities, we summarize each utility into a single vector by the self-attention mechanism, in which the summarized vector of question utility is denoted by $c_q$. We concatenate all these vectors together with $c_r$, projecting it into a $d$-dimensional vector of context representation $c$. 

The decoder architecture is similar to the generative decoder described in Sec. \ref{sec:gen_decoder} except that the input of the decoder at the $i$-th step is the concatenation of $w_{i-1}$, $c_q$, and $c_r$. At the time of inference, we use the beam search technique to efficiently find the most likely hypothesis generated by the decoder.

\subsection{Experimental Setup}
Following \cite{hori2019end}, we perform the experiment on the AVSD prototype which is split into training, validation, and test sets with 6172, 732, and 733 videos, respectively. Each video is collected from the Charades dataset, annotated with a caption and 10 dialog rounds.
The hidden size $d$ is set to 512; the GLoVe vectors are $300$-dimensional. We train the models in 15 epochs using the Adam optimizer with initial learning rate $1\times10^{-3}$ in all the experiments. The dropout with rate of 0.2 is applied for the LSTMs.

\subsection{Experimental Results}
   
Table \ref{tab:avsd} shows 
the results, which include evaluation on a number of metrics to measure the quality of generated answers, i.e. CIDEr, BLEU, METEOR, ROUGE\_L. It is seen that our models outperform the baselines presented in \cite{hori2019end} over all the metrics; {
specifically, it improves the CIDEr score by 22.3\% (from 0.618 to 0.841) with VGG features and by 12.4\% (from 0.727 to 0.851) with I3D features.}

\newcommand{\figw}{1.4\textwidth}

\begin{figure*}[h]
\centering
\centerline{
\includegraphics[width=\figw]{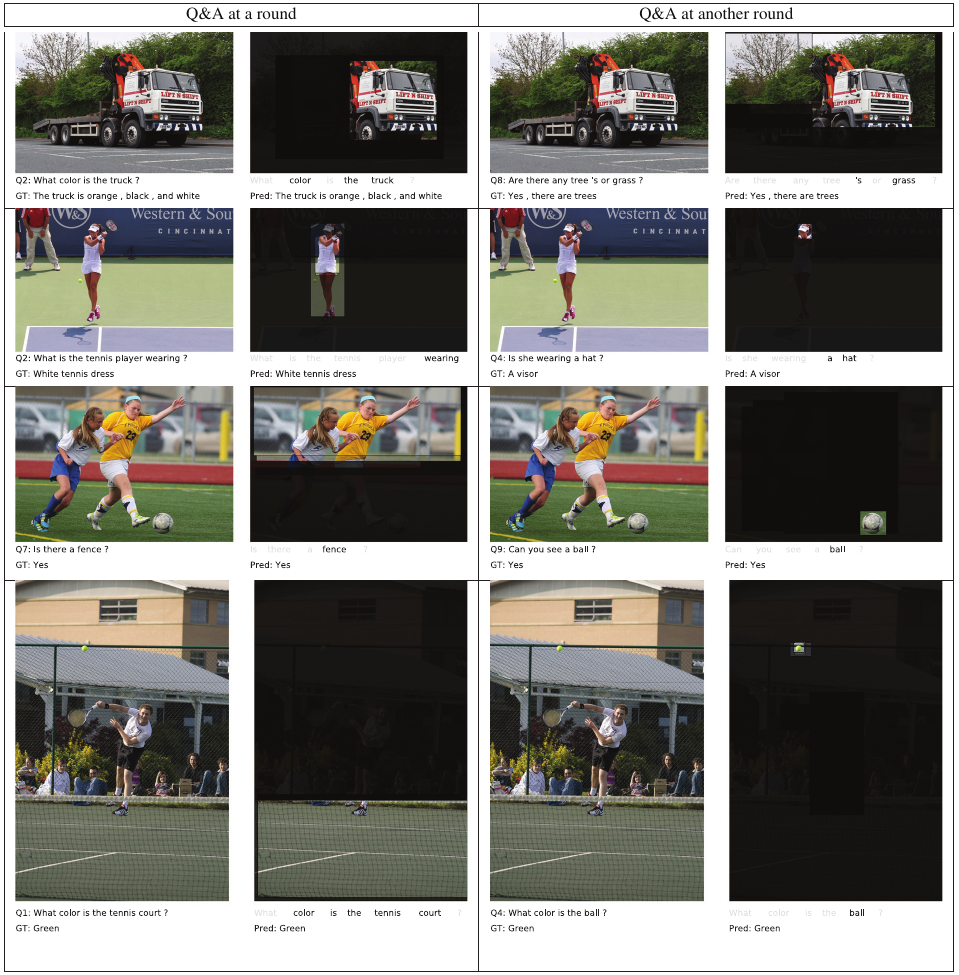}
}
\caption{Examples of results for which the top-1 prediction is the same as the ground truth answer on the validation split of Visdial v1.0. Each row shows selected two rounds of Q\&A for one image.}

\label{fig:s1}
\end{figure*}

\begin{figure*}[h]
\centering
\centerline{
\includegraphics[width=\figw]{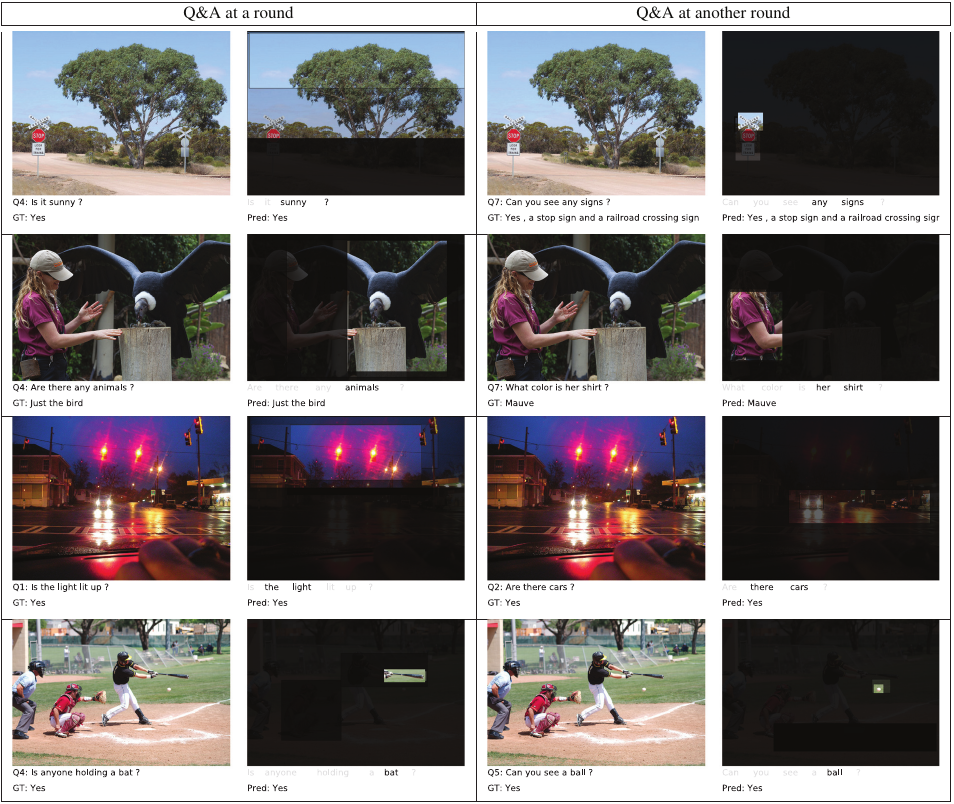}
}
\caption{Examples of results for which the top-1 prediction is the same as the ground truth answer on the validation split of Visdial v1.0. Each row shows selected two rounds of Q\&A for one image.}
\label{fig:s2}
\end{figure*}

\begin{figure*}[h]
\centering
\centerline{
\includegraphics[width=\figw]{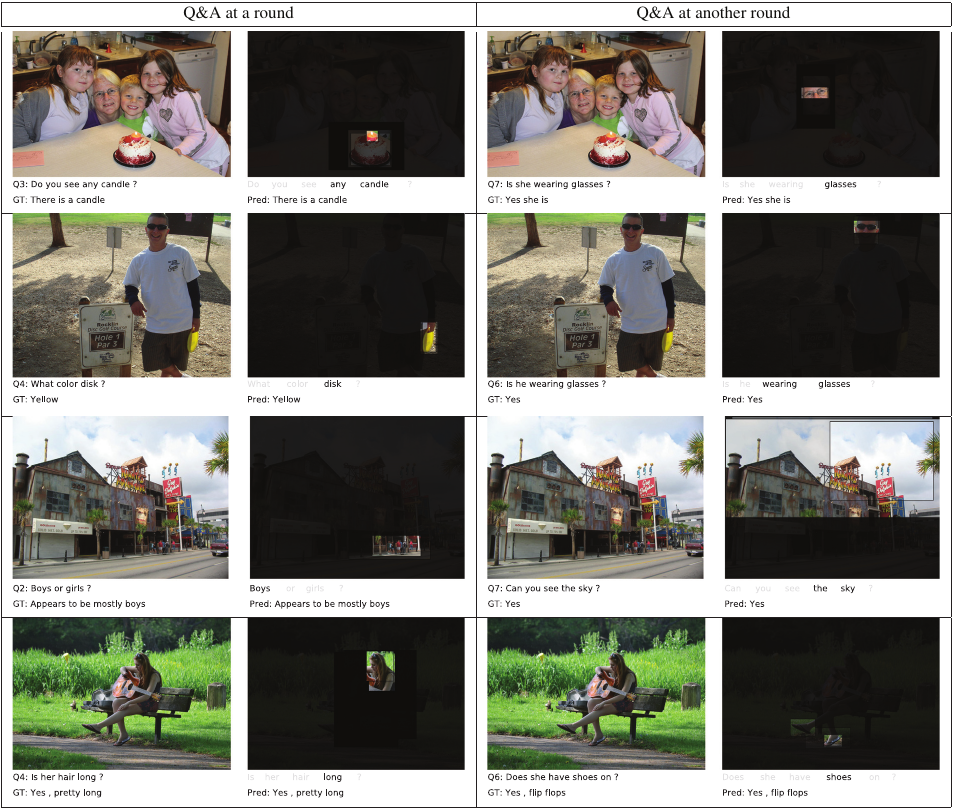}
}
\caption{Examples of results for which the top-1 prediction is the same as the ground truth answer on the validation split of Visdial v1.0. Each row shows selected two rounds of Q\&A for one image.}
\label{fig:s3}
\end{figure*}


\begin{figure*}[h]
\centering
\centerline{
\includegraphics[width=\figw]{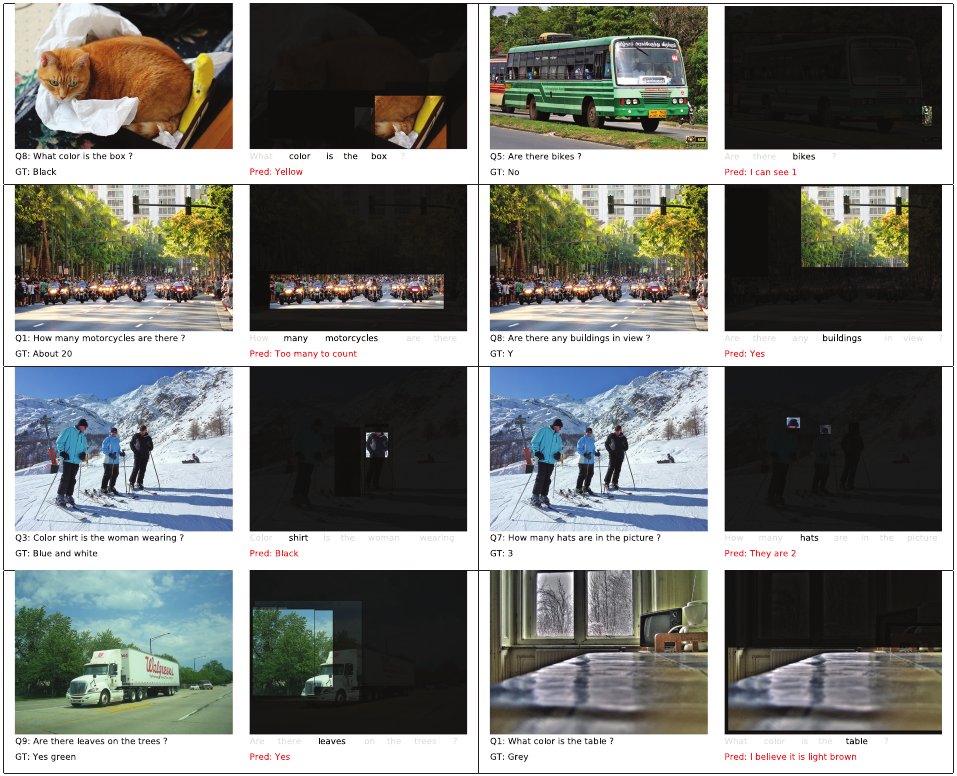}
}
\caption{
Examples of results for which the top-1 prediction is different from the ground truth answer on the validation split of Visdial v1.0.}
\label{fig:d2}
\end{figure*}

\begin{figure*}[h]
\centering
\centerline{
\includegraphics[width=\figw]{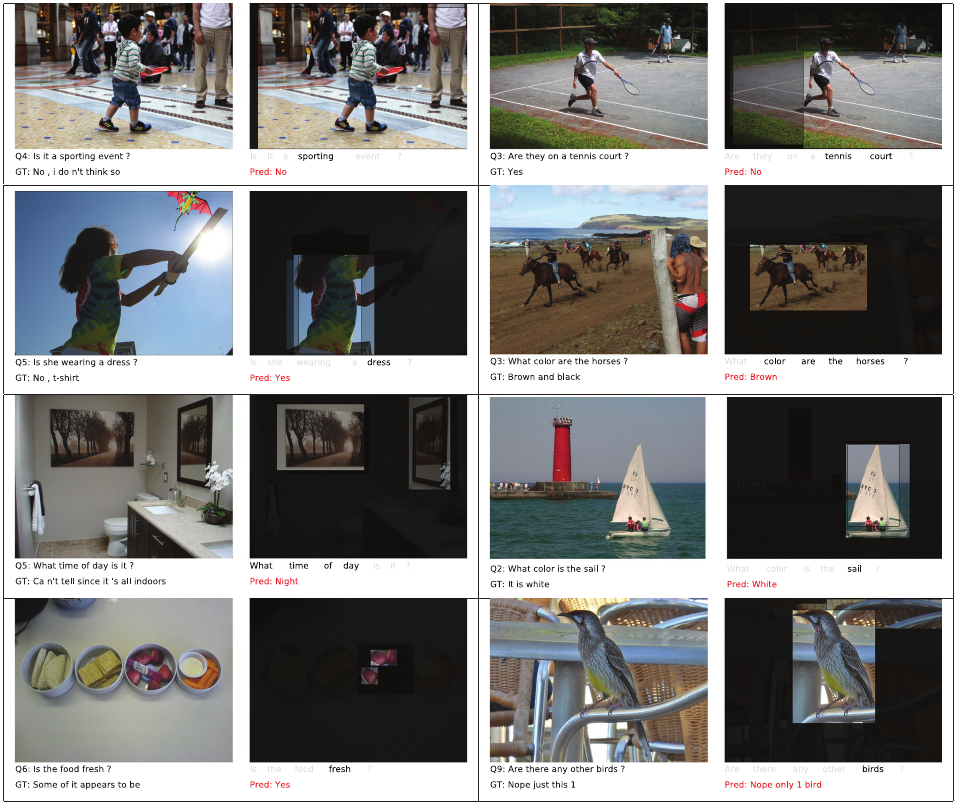}
}
\caption{
Examples of results for which the top-1 prediction is different from the ground truth answer on the validation split of Visdial v1.0.}
\label{fig:d3}
\end{figure*}

\end{document}